\pdfoutput=1

\documentclass[11pt]{article}

\usepackage{acl}

\usepackage{times}
\usepackage{latexsym}

\usepackage[T1]{fontenc}
\usepackage{algorithm}
\usepackage{algpseudocode}
\usepackage{graphicx}
\usepackage{amsmath}
\usepackage{amsfonts}
\usepackage{amssymb}
\usepackage{hyperref} 
\usepackage{booktabs} 
\usepackage{multirow}
\usepackage{colortbl}
\usepackage{makecell}
\usepackage{float}
\usepackage{enumitem}
\usepackage{xcolor} 
\usepackage[utf8]{inputenc}
\usepackage{ulem}
\usepackage{microtype}

\usepackage{inconsolata}

\title{Calibrating the Confidence of Large Language Models by Eliciting Fidelity}

\author{Mozhi Zhang\textsuperscript{1,2}\thanks{{} {} Work done during internship at Meituan.},
Mianqiu Huang\textsuperscript{1,2}, 
Rundong Shi\textsuperscript{3}, \\
{\bf Linsen Guo \textsuperscript{3}},
{\bf Chong Peng \textsuperscript{3}},
{\bf Peng Yan \textsuperscript{3}},
{\bf Yaqian Zhou\textsuperscript{1,2}}
{\bf Xipeng Qiu\textsuperscript{1,2}\thanks{{} {} Corresponding author.}} \\
  \textsuperscript{1}School of Computer Science, Fudan University \\
  \textsuperscript{2}Shanghai Key Laboratory of Intelligent Information Processing, Fudan University \\
  \textsuperscript{3} Meituan\\
  \texttt{\{mzzhang22, mqhuang23\}@m.fudan.edu.cn} \\
  \texttt{\{shirundong\}@meituan.com} \\
  \texttt{\{zhouyaqian, xpqiu\}@fudan.edu.cn} \\}

\begin{document}
\maketitle
\begin{abstract}
Large language models optimized with techniques like RLHF have achieved good alignment in being helpful and harmless. 
However, post-alignment, these language models often exhibit overconfidence, where the expressed confidence does not accurately calibrate with their correctness rate. 
In this paper, we decompose the language model confidence into the \textit{Uncertainty} about the question and the \textit{Fidelity} to the answer generated by language models. 
Then, we propose a plug-and-play method, \textit{UF Calibration}, to estimate the confidence of language models. 
Our method has shown good calibration performance by conducting experiments with 6 RLHF-LMs on four MCQA datasets. 
Moreover, we propose two novel metrics, IPR and CE, to evaluate the calibration of the model, and we have conducted a detailed discussion on \textit{Truly Well-Calibrated Confidence} for large language models.
Our method could serve as a strong baseline, and we hope that this work will provide some insights into the model confidence calibration. 
\end{abstract}

\section{Introduction}
Large language models (LLMs) acquire vast world knowledge and demonstrate powerful capabilities through pre-training~\cite{brown2020language, openai2023gpt4, bubeck2023sparks, Sun2024MOSS}. With technologies like RLHF~\cite{ouyang2022training} and RLAIF~\cite{bai2022constitutional, lee2023rlaif}, large language models can become more helpful and harmless to align with human preferences~\cite{askell2021general}.
However, how to build a more honest system has not yet been fully discussed. An honest model should have a certain understanding of the boundary of its knowledge, that is,~\textit{knowing what it does not know}~\cite{yin-etal-2023-large, yang2023alignment, zhou-etal-2024-open-world}.
A plausible method is utilizing the calibrated confidence to estimate the knowledge boundary of language models.
For pre-trained language models, the per-token logit can already be considered a well-calibrated confidence score, which implies that \textit{pre-trained language models (mostly) know what they know}~\cite{kadavath2022language}.

However, recent studies have indicated that language models optimized with techniques like RLHF will exhibit issues of overconfidence~\cite{lin2022teaching, kadavath2022language, openai2023gpt4, he2023investigating, zhao2023automatic,  tian-etal-2023-just, xiong2023llms}. 
This issue could be reflected in Multiple-Choice Question Answering (MCQA) tasks, where the probability of RLHF-LMs generating a token and the likelihood of that token being the correct answer are not well-calibrated.
For example, an answer provided by RLHF-LMs with 95\% confidence does not mean that there is a 95\% probability that the answer is correct. 
This phenomenon may be due to the optimization objective of RLHF, which is to make the model generate responses aligned with human preferences rather than fitting answers that appear more frequently in the corpus during the pre-training stage.
\begin{figure}[!t]
  \centering
  \includegraphics[width=0.75\columnwidth]{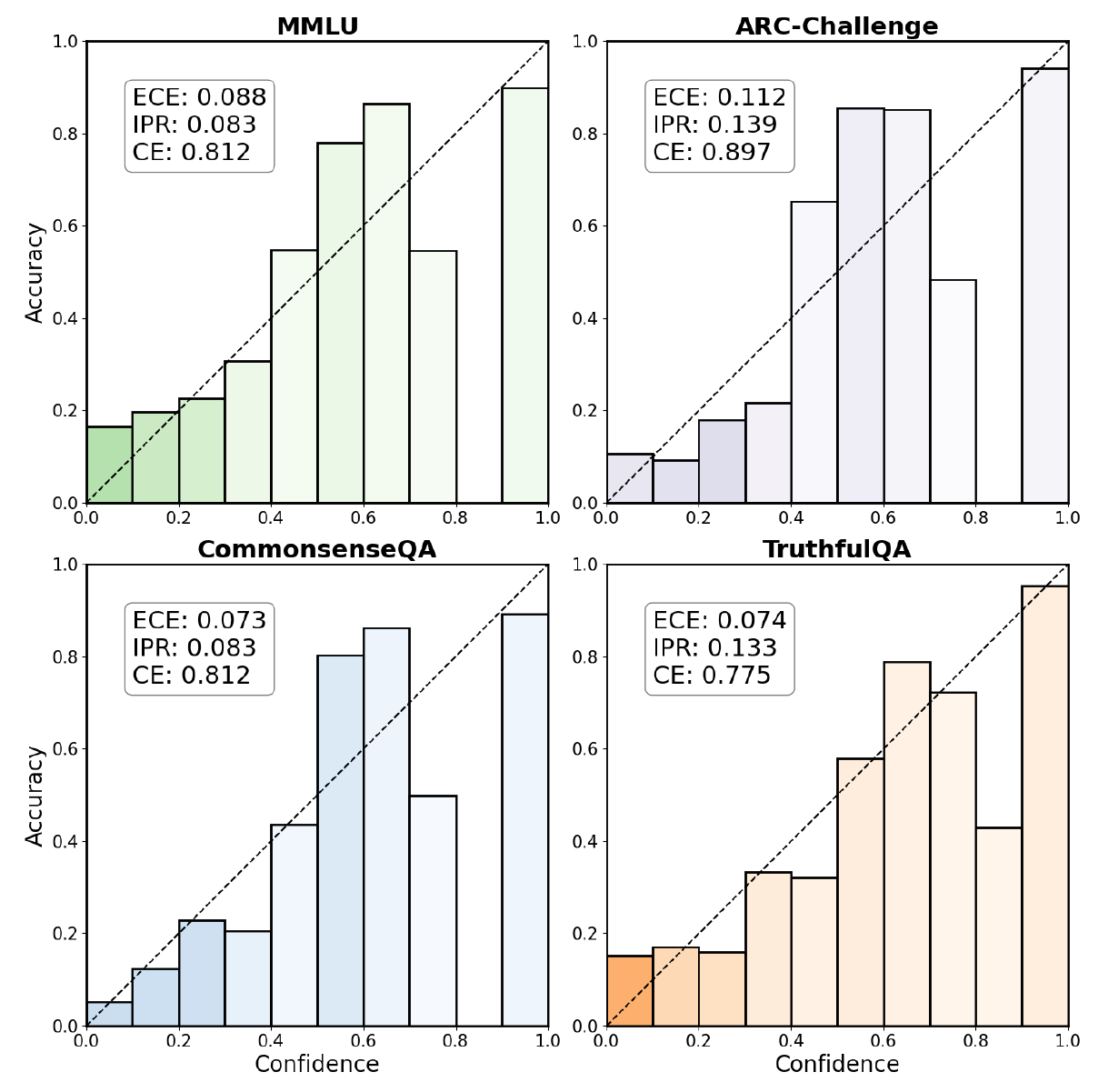}
  \caption{In four different MCQA datasets, our method has demonstrated good calibration effects, meaning it is sufficiently close to the $y=x$ curve. The experimental data is derived from \texttt{GPT-3.5-Turbo}.} \label{fig:show}
  \vspace{-15pt}
\end{figure}

To alleviate the issue of miscalibration, previous work focuses on two perspectives: the logit-based method and the verbalization-based method.
Logit-based methods are usually post-hoc. We need to find a higher temperature (usually above 2.0), known as Temperature-Tuning~\cite{guo2017calibration}, to make the distribution of the model's token logit smoother for mitigating overconfidence~\cite{kadavath2022language, he2023investigating}.
The verbalization-based method usually requires prompt engineering to elicit the model's confidence, and it also necessitates the model to have strong Self-Awareness~\cite{lin2022teaching, tian-etal-2023-just, yin-etal-2023-large}.
Aggregating the model's logit-based and verbalization-based confidence can also calibrate the model confidence to some extent~\cite{xiong2023llms}.

As shown in Figure~\ref{fig:hot} and Appendix Tabel~\ref{table: whyfidelity}, by replacing the model's answer with ``\textit{All other options are wrong.}'', we can assess whether the model had high fidelity to its previously given answer.
Inspired by this phenomenon, we decompose the language model confidence into two dimensions: the \textbf{\textit{Uncertainty}} about the question and the \textbf{\textit{Fidelity}} to the answer generated by language models.
First, if the answers provided by language model are consistent under multiple samplings, it indicates that language model has lower uncertainty regarding that question.
Thus, we could utilize the information entropy of the frequency distribution of sampled answers to calculate the model's uncertainty about a question.
Second, we design a novel method to estimate the model's fidelity to each of its sampled answers.
Last, the uncertainty regarding question $\mathcal{Q}$ and the fidelity to the answer $a_i$ together determine the model's confidence.
As shown in Figure~\ref{fig:show}, our proposed UF Calibration achieved good calibration across different MCQA datasets.
Meanwhile, UF Calibration does not require knowledge of the model's per-token log-probability, making it broadly applicable to various Black-box RLHF-LMs, which do not provide the per-token log-probability.

To have a closer look at the calibration of model confidence, we propose two novel metrics for evaluating and observation:
\textbf{1)}~\textit{\textbf{I}nverse \textbf{P}air \textbf{R}atio}~(IPR), which is the proportion of inverse pairs in the Reliability Diagram. 
This metric could reflect whether the model is well-calibrated from the perspective of the monotonicity of the Reliability Diagram.
If the reliability diagram is monotonic, it indicates that the average accuracy of low-confidence answers is always lower than that of high-confidence answers.
\textbf{2)} As shown in Table~\ref{table: whyCE}, we find that as the number of model parameters increases, language models still tend to consistently express uncertainty within certain fixed ranges.
Thus, we design the~\textit{\textbf{C}onfidence \textbf{E}venness}~(CE) to observe to the uniformity of the density of each bar in the reliability diagram.
Our experimental results indicate that, after calibration, even within the same dataset, there is a significant difference in the confidence of the answers provided by language models for different questions. 
We summarize our main contributions as follows:
\vspace{-0.5em}
\begin{enumerate}[label={\arabic*)}, leftmargin=*]
\item Our proposed method could be viewed as a strong baseline for eliciting model confidence, where answer set is known. And the calibrated confidence could be viewed as a soft label.
\vspace{-0.5em}
\item We propose two new metrics, $\mathrm{IPR}$ and $\mathrm{CE}$, to evaluate the calibration of LM's confidence.
\vspace{-0.5em}
\item We conduct a detailed discussion of a research question: ``\textit{What kind of Confidence is Truly Well-Calibrated?}'', and we hope our discussion can bring some insights to the community.
\end{enumerate}

\section{Related Work}
Recent work has focused on LLM calibration~\cite{lin2022teaching, kadavath2022language, openai2023gpt4}. 
In this section, we will briefly introduce two mainstream methods for eliciting the confidence from language models, namely the Logit-based Method and the Verbalization-based Method.
\subsection{Logit-based Method}
When we can obtain the per-token logits from language models, we can directly use the probability of generating candidate answers as its confidence.
\begin{align}
    \mathrm{Conf}(a_i) = \frac{\exp(\mathrm{logit}_{a_i} / t)}{\sum_{j=1}^{|\mathcal{A}|}\exp(\mathrm{logit}_{a_j} / t)}, \label{ali: logit-based method}
\end{align} 
where $t$ is the sampling temperature of language models and $|\mathcal{A}|$ is the size of candidate answer set $\mathcal{A}$.
Recent studies indicate that good calibration can be achieved by adjusting the temperature of RLHF-LMs~\cite{kadavath2022language, he2023investigating}. However, temperature-scaling~\cite{guo2017calibration} often requires higher temperatures, such as above 2.0~\cite{kadavath2022language}, which might cause the outputs of the language models to become too random. 
When the probabilities for model-generated tokens are inaccessible, a straightforward solution is to deploy sampling and use the frequency of the sampled result to estimate the probability of generating this token. 
For instance, given a question $\mathcal{Q}$, we could sample $K$ times to acquire a set of answers $\mathcal{A}$ containing $N$ distinct answers, and each answer with an associated frequency $n_i$. The probability of the model generating answer $a_i$ can be estimated by $\frac{n_i}{K}$. 
Therefore, we could estimate the confidence of language models by $\mathcal{P}_{\mathrm{sampled}}(a_i)$. 
Recently, ~\citet{kumar2023conformal} also propose to utilize the conformal prediction to calibrate the confidence of LLMs.
\begin{align}
    \mathrm{Conf}(a_i) = \mathcal{P}_{\mathrm{sampled}}(a_i) = \frac{n_i}{K}, a_i \in \mathcal{A}
\end{align}

\begin{figure}[!t]
  \centering
  \includegraphics[width=0.75\columnwidth]{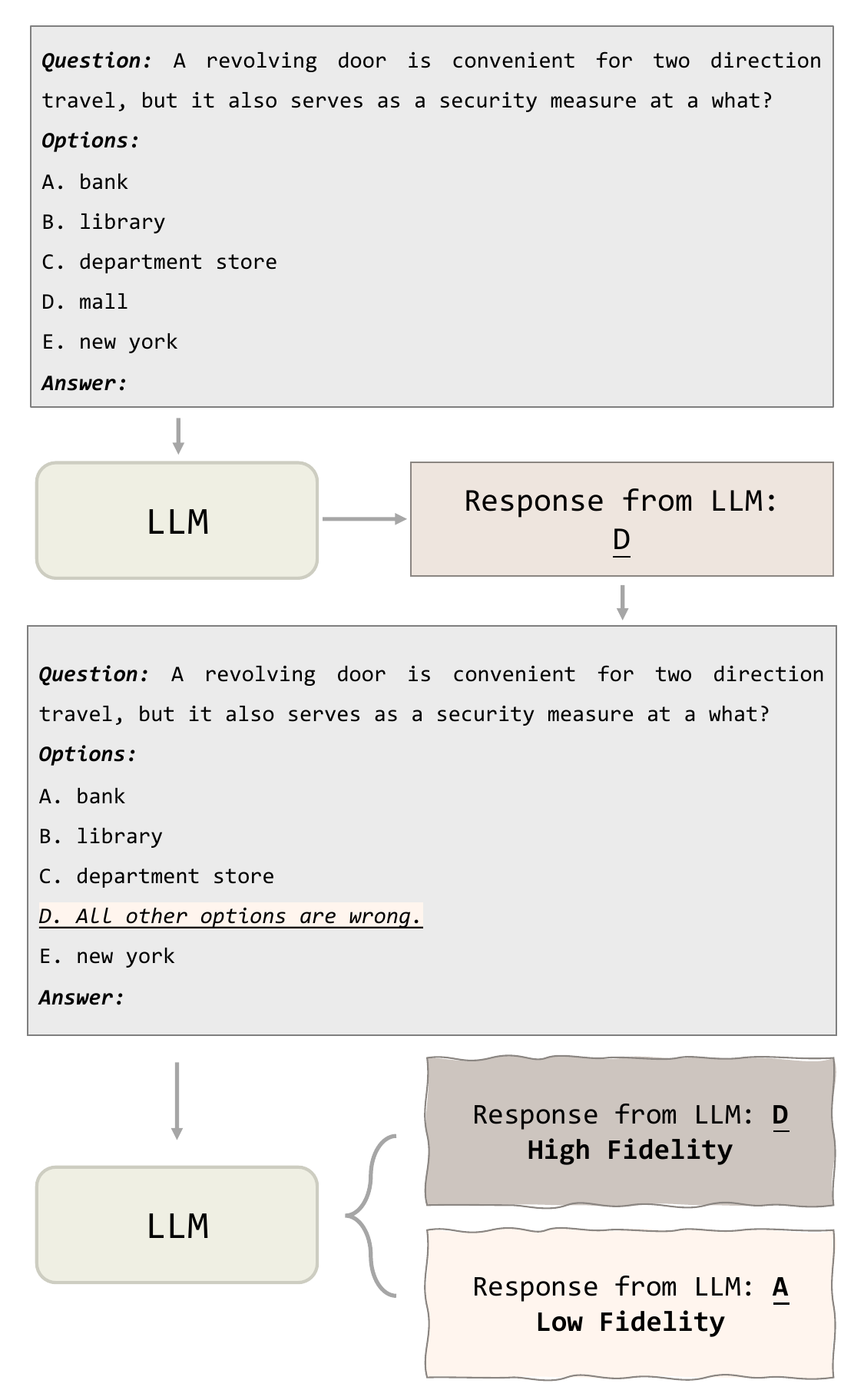}
  \vspace{-0.3em}
  \caption{If the model's choice of answer changes after replacing the content of its previous selected option with ``\textit{All other options are wrong}'', it could be considered that the model's fidelity to its previous answer is not high enough.} \label{fig:hot}
  \vspace{-10pt}
\end{figure}

\subsection{Verbalization-based Method}
However, some commercial models, such as ChatGPT and Claude, usually do not provide per-token logits. 
Benefiting from instruction fine-tuning\cite{chung2022scaling, zhang2023instruction}, language models could generate responses corresponding to the input instructions.
Another intuitive method is to prompt large language models to  provide their verbalized confidence along with their responses as follows~\cite{jiang-etal-2021-know, lin2022teaching, tian-etal-2023-just}:
\begin{align}
    (\mathrm{Answer},~\mathrm{Conf}) = \mathrm{LLM}(\mathrm{Question}),
\end{align}
This method requires the model to have a strong ability to follow instructions and strong self-awareness (know whether it knows something or not~\cite{yin-etal-2023-large}).
Accordingly, verbalized confidence can be a floating-point number between 0 and 1, i.e., \texttt{`0.8'}. And it can be linguistic expressions, i.e., \texttt{`Almost Certain'}, \texttt{`About Even'}, \texttt{`Unlikely'}.
Although this method is quite easy to implement, we find various different LMs always tend to output some fixed high confidence expressions, as show in Table~\ref{table: whyCE}.
\section{Methodology}
In this section, we will introduce the method we propose. 
Our method does not require any knowledge of the per-token logit of language models or trivial prompt engineering to make the language model output its confidence in a specified format. 

\begin{figure*}[!t]
  \centering
  \includegraphics[width=0.85\textwidth]{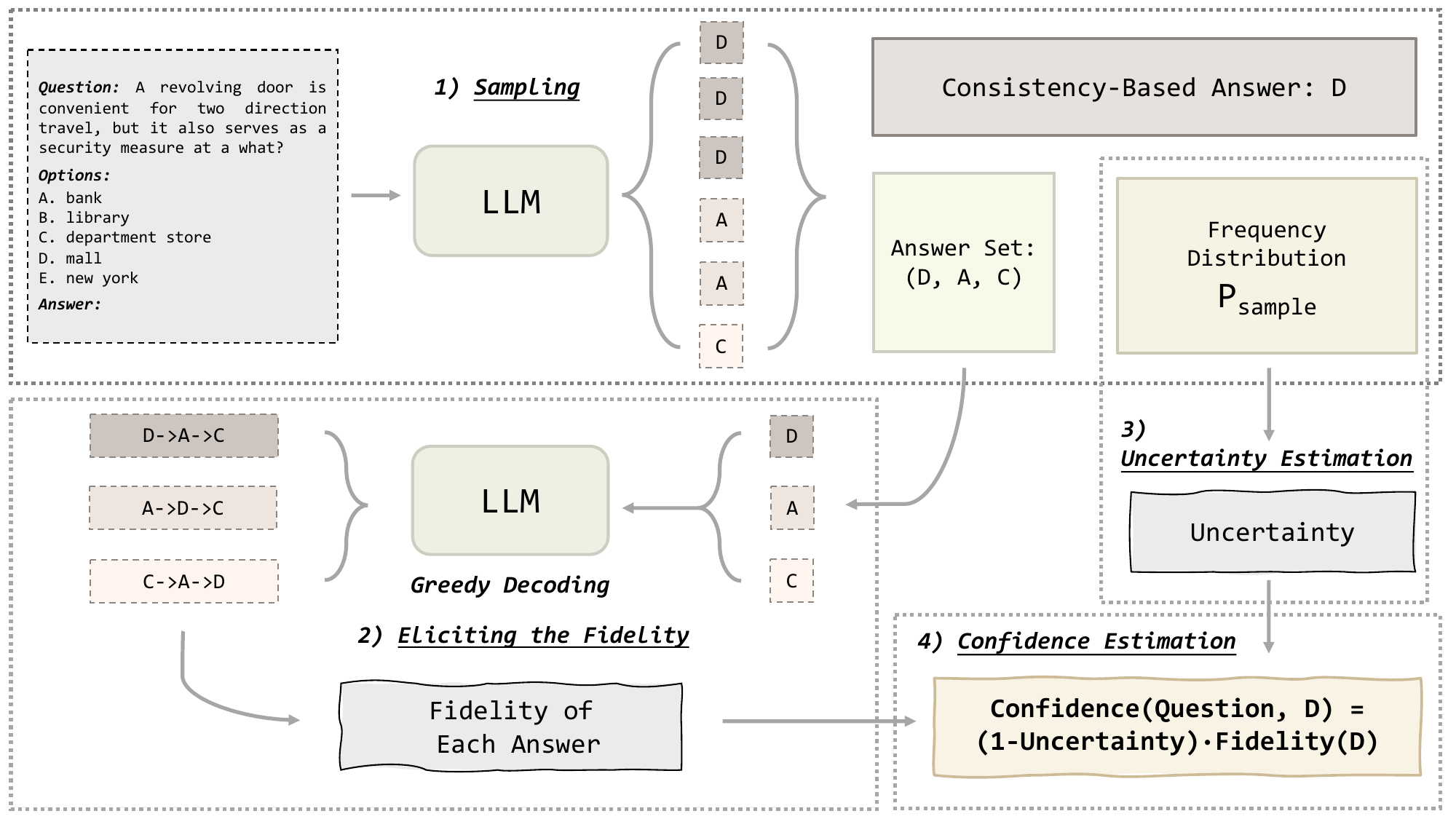}
  \caption{Our proposed UF Calibration, which requires at most two phases to invoke the model. In the Sampling phase, for black-box models, similar to the Sampled method, we need to sample 10 times. For white-box models, a single invocation is sufficient. In the eliciting the fidelity phase, the model needs to be invoked approximately 2 to 3 times to generate a fidelity chain, as show in Table~\ref{table: averageLength}.} \label{fig:framework}
\end{figure*}

\subsection{Sampling}\label{sec:method_sampling}
Firstly, as shown in the first step from Figure \ref{fig:framework}, for question $\mathcal{Q}$, by sampling $K$ times, we can obtain a set of candidate answers $\mathcal{A}$.
We take the most frequently occurring answer as the final answer.
Meanwhile, we can obtain the frequency distribution $\mathcal{P}_{\mathrm{sampled}}$ of candidate answers. 

\subsection{Eliciting the Fidelity of Answers}\label{sec:method_fidelity}

As shown in Figure \ref{fig:hot}, for question $\mathcal{Q}$ and a candidate answer ($a_i$, $o_i$), where the option index is $a_i$ and the content is $o_i$, we simply replace $o_i$ with ``\textbf{\textit{All other options are wrong.}}'', and then query the model again.
If the model has high fidelity to the previously selected answer ($a_i$, $o_i$), it should select ($a_i$, ``\textit{All other options are wrong.}'') in the subsequent round of inquiry rather than any other option.
If language models select other options, we remove the newly selected option to ensure that there is only one ``\textit{All other options are wrong}'' in candidate options.
By repeating this process until the model selects ``\textit{All other options are wrong}'', we can establish a hierarchical fidelity chain $\mathcal{C}$, such as "A$\rightarrow$C$\rightarrow$D".
This implies that when all options are available, the model will prefer to select option A. 
However, if option A is excluded, the model will tend to choose option C, which indicates that the model's fidelity to option A is not high enough.
Accordingly, if the chain $\mathcal{C}$ has only one element, such as ``A'', this suggests that the model's fidelity to option A is high enough, which can, to a certain extent, reflect the model's confidence.
Correspondingly, for a hierarchical fidelity chain  $\mathcal{C}$, we assign a fidelity weight to each element from right to left. 
For example, for the \(i\)th element $d_i$ from the right, we simply set its weight as $\tau^i$. 
Therefore, the normalized fidelity of the \(i\)th element $a_i$ can be calculated as follows:
\begin{align}
  \mathbf{Fidelity}_{\mathcal{C}}(a_i) = \frac{\tau^i}{\sum_{i=1}^{|\mathcal{C}|} \tau^i}, \label{ali: fidelity}
\end{align} 
where we usually set $\tau$ as 2.
As shown in Figure~\ref{fig:framework}, the answer set $\mathcal{A}$ might include multiple different answers. 
Consequently, we sequentially replace the candidate answer in $\mathcal{A}$ with ``\textit{All other options are wrong.}'' to elicit different hierarchical fidelity chains, as depicted in the second step of Figure~\ref{fig:framework}. 
The fidelity score of each element $a_i$ in every hierarchical fidelity chain $\mathcal{C}_j$ can be calculated using~(\ref{ali: fidelity}). 
Thus, the model's fidelity of answer $a_{i}$ can be calculated by the weighted average fidelity score across different hierarchical chains. 
Since the hierarchical fidelity chain is elicited by greedy decoding, the frequency of occurrence of different chains is consistent with the frequency of occurrence of the first element $a_{|\mathcal{C}|}$ from left to right. 
Therefore, the frequency $\mathcal{P}_{\mathrm{sampled}}(a_{|\mathcal{C}|})$ can be viewed as a proxy for the probability  $\mathcal{P}_{\mathrm{sampled}}({\mathcal{C}_j})$ of different hierarchical fidelity chains to calculate the overall fidelity score $\mathbf{F}(\cdot)$ of each answer.
\begin{align}
    \mathbf{F}(a_i) = \sum_{j=1}^{|\mathcal{A}|} \mathcal{P}_{\mathrm{sampled}}({\mathcal{C}_j})\cdot \mathbf{Fidelity}_{\mathcal{C}_j}(a_i),
\end{align}

\subsection{Uncertainty Estimation}\label{sec:Uncertainty_Estimation}
As shown in Section \ref{sec:method_sampling}, through sampling, we can obtain the frequency of each answer generated by the model and use it to estimate the generation probability of each answer token. 
Previous works~\cite{kadavath2022language, openai2023gpt4} have revealed that RLHF-LMs often exhibit overconfidence in token generation probability, especially in the temperature range we commonly use, such as between 0 and 1.0.
However, these probabilities could still reveal, to some extent, the model's confidence regarding the current question $\mathcal{Q}$. 
For instance, if the distribution of $\mathcal{P}_{\mathrm{sampled}}$ is flatter, it indicates that the language model has more significant uncertainty regarding the question $\mathcal{Q}$.
An intuitive method is calculating the information entropy of the distribution $\mathcal{P}_{\mathrm{sampled}}$ to  estimate the model's uncertainty about question $\mathcal{Q}$ as follows:
\begin{align}
    \mathbf{Uncertainty}(\mathcal{Q}) = - \frac{\sum^{\mathrm{M}}_{i=1}p_i\cdot\log p_i}{\log\mathrm{M}}, \label{ali: uncertainty}
\end{align}
where $\mathrm{M}$ is the option number of question $\mathcal{Q}$. Since the range of the information entropy for $\mathcal{P}_{\mathrm{sampled}}$ is from 0 to $\log\mathrm{M}$, we normalize the information entropy using $\log\mathrm{M}$.

\definecolor{top1color}{HTML}{AECDD7} 
\definecolor{top2color}{HTML}{E3EEEF} 
\definecolor{top3color}{HTML}{FAE7D9} 
\definecolor{top4color}{HTML}{F0B79A} 


\begin{table*}[!ht]
\centering
\small
\resizebox{\textwidth}{!}{\begin{tabular}{lccccccccccccccccc}
\toprule
& \multicolumn{4}{c}{\textbf{ARC-Challenge}} &  \multicolumn{4}{c}{\textbf{MMLU}} & \multicolumn{4}{c}{\textbf{CommonSenseQA}} &  \multicolumn{4}{c}{\textbf{TruthfulQA}} \\
\cmidrule(lr){2-5} \cmidrule(lr){6-9} \cmidrule(lr){10-13} \cmidrule(lr){14-17}
\textbf{Method} 
& \makecell{$\mathrm{ECE}_{10}$ \scalebox{0.65}{$\downarrow$}} & \makecell{$\mathrm{IPR}_{10}$ \scalebox{0.65}{$\downarrow$}} & \makecell{$\mathrm{CE}_{10}$ \scalebox{0.65}{$\uparrow$}} & \makecell{\textbf{Acc} \scalebox{0.65}{$\uparrow$}}  & 
 \makecell{$\mathrm{ECE}_{10}$ \scalebox{0.65}{$\downarrow$}} & \makecell{$\mathrm{IPR}_{10}$ \scalebox{0.65}{$\downarrow$}} & \makecell{$\mathrm{CE}_{10}$ \scalebox{0.65}{$\uparrow$}} & \makecell{\textbf{Acc} \scalebox{0.65}{$\uparrow$}}  & 
 \makecell{$\mathrm{ECE}_{10}$ \scalebox{0.65}{$\downarrow$}} & \makecell{$\mathrm{IPR}_{10}$ \scalebox{0.65}{$\downarrow$}}& \makecell{$\mathrm{CE}_{10}$ \scalebox{0.65}{$\uparrow$}}  & \makecell{\textbf{Acc} \scalebox{0.65}{$\uparrow$}} &  \makecell{$\mathrm{ECE}_{10}$ \scalebox{0.65}{$\downarrow$}} & \makecell{$\mathrm{IPR}_{10}$ \scalebox{0.65}{$\downarrow$}} & \makecell{$\mathrm{CE}_{10}$ \scalebox{0.65}{$\uparrow$}} & \makecell{\textbf{Acc} \scalebox{0.65}{$\uparrow$}} &  \\ 
\midrule
\multicolumn{17}{c}{\textsc{GPT-3.5-Turbo}} \\ 
\midrule
Verb
& \cellcolor{top1color}\textbf{0.069} & \cellcolor{top3color}0.200  & \cellcolor{top3color}0.681  & \cellcolor{top4color}75.597 
& \cellcolor{top3color}0.138 & \cellcolor{top3color}0.200  & \cellcolor{top3color}0.795  & \cellcolor{top3color}59.028
& \cellcolor{top2color}\uline{0.087} & \cellcolor{top3color}0.178  & \cellcolor{top3color}0.660  & \cellcolor{top4color}71.253
& \cellcolor{top3color}0.215 & \cellcolor{top3color}0.178  & \cellcolor{top2color}\uline{0.792}  & \cellcolor{top4color}57.405  \\
Ling
& \cellcolor{top2color}\uline{0.083} & \cellcolor{top4color}0.464  & \cellcolor{top4color}0.451  & \cellcolor{top3color}75.683
& \cellcolor{top4color}0.197 & \cellcolor{top4color}0.472  & \cellcolor{top4color}0.441  & \cellcolor{top4color}56.019 
& \cellcolor{top3color}0.109 & \cellcolor{top4color}0.250  & \cellcolor{top4color}0.451  & \cellcolor{top3color}71.499
& \cellcolor{top4color}0.271 & \cellcolor{top4color}0.667  & \cellcolor{top4color}0.669  & \cellcolor{top3color}59.241 \\
Sampled
& \cellcolor{top3color}0.095 & \cellcolor{top1color}\textbf{0.067}  & \cellcolor{top2color}\uline{0.793}  & \cellcolor{top1color}\textbf{79.266}
& \cellcolor{top2color}\uline{0.120} & \cellcolor{top1color}\textbf{0.022}  & \cellcolor{top1color}\textbf{0.922}  & \cellcolor{top1color}\textbf{63.151}  
& \cellcolor{top4color}0.135 & \cellcolor{top1color}\textbf{0.067}  & \cellcolor{top2color}0.782  & \cellcolor{top1color}\textbf{74.590}
& \cellcolor{top2color}\uline{0.147} & \cellcolor{top1color}\textbf{0.044}  & \cellcolor{top1color}\textbf{0.901}  & \cellcolor{top1color}\textbf{59.333} \\
\midrule
Ours
& \cellcolor{top4color}0.112 & \cellcolor{top2color}\uline{0.139}  & \cellcolor{top1color}\textbf{0.897}  & \cellcolor{top1color}\textbf{79.266} 
& \cellcolor{top1color}\textbf{0.088} & \cellcolor{top2color}\uline{0.083}  & \cellcolor{top2color}\uline{0.812}  & \cellcolor{top1color}\textbf{63.151} 
& \cellcolor{top1color}\textbf{0.073} & \cellcolor{top2color}\uline{0.083}  & \cellcolor{top1color}\textbf{0.812}  & \cellcolor{top1color}\textbf{74.590}
& \cellcolor{top1color}\textbf{0.074} & \cellcolor{top2color}\uline{0.133}  & \cellcolor{top3color}0.775  & \cellcolor{top1color}\textbf{59.333}  \\
\midrule
\multicolumn{17}{c}{\textsc{GPT-4-Turbo}} \\ 
\midrule
Verb
& \cellcolor{top3color}0.080 & \cellcolor{top4color}0.400  & \cellcolor{top2color}\uline{0.642}  & \cellcolor{top1color}\textbf{92.833}   
& \cellcolor{top1color}\textbf{0.045} & \cellcolor{top3color}0.095  & \cellcolor{top2color}\uline{0.706}& \cellcolor{top1color}\textbf{81.25}
& \cellcolor{top2color}\uline{0.083} & \cellcolor{top3color}0.111  & \cellcolor{top2color}\uline{0.713}  & \cellcolor{top4color}83.210 
& \cellcolor{top2color}\uline{0.056} & \cellcolor{top1color}\textbf{0.044}  & \cellcolor{top3color}0.598  & \cellcolor{top3color}83.109    \\
Ling
& \cellcolor{top1color}\textbf{0.040} & \cellcolor{top1color}\textbf{0.036}  & \cellcolor{top3color}0.520  & \cellcolor{top4color}89.505 
& \cellcolor{top2color}\uline{0.066} & \cellcolor{top1color}\textbf{0.083}  & \cellcolor{top3color}0.627  & \cellcolor{top4color}78.762  
& \cellcolor{top1color}\textbf{0.056} & \cellcolor{top1color}\textbf{0.071}  & \cellcolor{top3color}0.637  & \cellcolor{top1color}\textbf{83.702}
& \cellcolor{top3color}0.059 & \cellcolor{top2color}\uline{0.139}  & \cellcolor{top2color}\uline{0.635}  & \cellcolor{top4color}79.437  \\
Sampled
& \cellcolor{top2color}\uline{0.067} & \cellcolor{top3color}0.200  & \cellcolor{top4color}0.221  & \cellcolor{top1color}\textbf{92.833} 
& \cellcolor{top4color}0.153 & \cellcolor{top4color}0.311  & \cellcolor{top4color}0.536  & \cellcolor{top2color}\uline{80.324}   
& \cellcolor{top4color}0.121 & \cellcolor{top4color}0.133  & \cellcolor{top4color}0.541  & \cellcolor{top2color}\uline{83.866} 
& \cellcolor{top4color}0.091 & \cellcolor{top4color}0.178  & \cellcolor{top4color}0.478  & \cellcolor{top1color}\textbf{87.515}  \\
\midrule
Ours
& \cellcolor{top4color}0.127 & \cellcolor{top2color}\uline{0.083}  & \cellcolor{top1color}\textbf{0.757}  & \cellcolor{top1color}\textbf{92.833} 
& \cellcolor{top3color}0.089 & \cellcolor{top1color}\textbf{0.083}  & \cellcolor{top1color}\textbf{0.906}  & \cellcolor{top2color}\uline{80.324}   
& \cellcolor{top3color}0.109 & \cellcolor{top2color}\uline{0.083}  & \cellcolor{top1color}\textbf{0.925}  & \cellcolor{top2color}\uline{83.866} 
& \cellcolor{top1color}\textbf{0.042} & \cellcolor{top1color}\textbf{0.044}  & \cellcolor{top1color}\textbf{0.764}  & \cellcolor{top1color}\textbf{87.515}   \\
\bottomrule
\end{tabular}}
\caption{Experimental results derived from \texttt{GPT-3.5-Turbo} and \texttt{GPT-4-Turbo}. For each column in the table, the closer the color is to blue, the better the calibration. And the closer it is to orange, the worse the performance. We also have bolded the best results, and for the second-best results, we have added an underline beneath them.}
\vspace{-3mm}
\label{table: chatgpt}
\end{table*}

\definecolor{top1color}{HTML}{AECDD7} 
\definecolor{top2color}{HTML}{E3EEEF}
\definecolor{top3color}{HTML}{F0F8FF}
\definecolor{top4color}{HTML}{FAE7D9} 
\definecolor{top5color}{HTML}{F0B79A}



\begin{table*}[!ht]
\centering
\small
\resizebox{\textwidth}{!}{\begin{tabular}{lccccccccccccccccc}
\toprule
& \multicolumn{4}{c}{\textbf{ARC-Challenge}} &  \multicolumn{4}{c}{\textbf{MMLU}} & \multicolumn{4}{c}{\textbf{CommonSenseQA}} &  \multicolumn{4}{c}{\textbf{TruthfulQA}} \\
\cmidrule(lr){2-5} \cmidrule(lr){6-9} \cmidrule(lr){10-13} \cmidrule(lr){14-17}
\textbf{Method} 
& \makecell{$\mathrm{ECE}_{10}$ \scalebox{0.65}{$\downarrow$}} & \makecell{$\mathrm{IPR}_{10}$ \scalebox{0.65}{$\downarrow$}} & \makecell{$\mathrm{CE}_{10}$ \scalebox{0.65}{$\uparrow$}} & \makecell{\textbf{Acc} \scalebox{0.65}{$\uparrow$}}  & 
 \makecell{$\mathrm{ECE}_{10}$ \scalebox{0.65}{$\downarrow$}} & \makecell{$\mathrm{IPR}_{10}$ \scalebox{0.65}{$\downarrow$}} & \makecell{$\mathrm{CE}_{10}$ \scalebox{0.65}{$\uparrow$}} & \makecell{\textbf{Acc} \scalebox{0.65}{$\uparrow$}}  & 
 \makecell{$\mathrm{ECE}_{10}$ \scalebox{0.65}{$\downarrow$}} & \makecell{$\mathrm{IPR}_{10}$ \scalebox{0.65}{$\downarrow$}}& \makecell{$\mathrm{CE}_{10}$ \scalebox{0.65}{$\uparrow$}}  & \makecell{\textbf{Acc} \scalebox{0.65}{$\uparrow$}} &  \makecell{$\mathrm{ECE}_{10}$ \scalebox{0.65}{$\downarrow$}} & \makecell{$\mathrm{IPR}_{10}$ \scalebox{0.65}{$\downarrow$}} & \makecell{$\mathrm{CE}_{10}$ \scalebox{0.65}{$\uparrow$}} & \makecell{\textbf{Acc} \scalebox{0.65}{$\uparrow$}} &  \\ 
\midrule
Verb.
& \cellcolor{top4color}0.135  & \cellcolor{top4color}0.178  & \cellcolor{top3color}0.752   & \cellcolor{top4color}58.191 
& \cellcolor{top4color}0.199  & \cellcolor{top4color}0.178  & \cellcolor{top3color}0.802   & \cellcolor{top4color}45.891 
& \cellcolor{top4color}0.107  & \cellcolor{top4color}0.083  & \cellcolor{top3color}0.806   & \cellcolor{top5color}59.214 
& \cellcolor{top4color}0.373  & \cellcolor{top2color}\uline{0.133}  & \cellcolor{top2color}\uline{0.874}   & \cellcolor{top4color}26.928  \\
Ling
& \cellcolor{top5color}0.298  & \cellcolor{top5color}0.286  & \cellcolor{top4color}0.613   & \cellcolor{top5color}50.853 
& \cellcolor{top5color}0.399  & \cellcolor{top5color}0.333  & \cellcolor{top4color}0.709   & \cellcolor{top5color}30.921 
& \cellcolor{top3color}0.097  & \cellcolor{top5color}0.222  & \cellcolor{top4color}0.771   & \cellcolor{top4color}60.770 
& \cellcolor{top5color}0.594  & \cellcolor{top5color}0.571  & \cellcolor{top4color}0.681   & \cellcolor{top5color}23.990  \\
Sampled
& \cellcolor{top3color}0.121  & \cellcolor{top2color}\uline{0.044}  & \cellcolor{top1color}\textbf{0.890}   & \cellcolor{top1color}\textbf{67.702}
& \cellcolor{top3color}0.162  & \cellcolor{top2color}\uline{0.067}  & \cellcolor{top1color}\textbf{0.919}   & \cellcolor{top2color}\uline{52.315} 
& \cellcolor{top5color}0.110  & \cellcolor{top2color}\uline{0.044}  & \cellcolor{top2color}\uline{0.857}   & \cellcolor{top2color}\uline{70.762}
& \cellcolor{top3color}0.236  & \cellcolor{top2color}\uline{0.133}  & \cellcolor{top1color}\textbf{0.891}   & \cellcolor{top2color}\uline{34.517}  \\
Token
& \cellcolor{top2color}\uline{0.064}  & \cellcolor{top3color}0.067  & \cellcolor{top5color}0.521   & \cellcolor{top3color}67.235
& \cellcolor{top2color}\uline{0.135}  & \cellcolor{top2color}\uline{0.067}  & \cellcolor{top5color}0.647   & \cellcolor{top1color}\textbf{54.803}
& \cellcolor{top2color}\uline{0.064}  & \cellcolor{top1color}\textbf{0.022}  & \cellcolor{top5color}0.477   & \cellcolor{top1color}\textbf{71.007} 
& \cellcolor{top2color}\uline{0.176}  & \cellcolor{top2color}\uline{0.133}  & \cellcolor{top5color}0.577   & \cellcolor{top1color}\textbf{34.761} \\
\midrule
Ours
& \cellcolor{top1color}\textbf{0.063}  & \cellcolor{top1color}\textbf{0.028}  & \cellcolor{top2color}\uline{0.887}   & \cellcolor{top1color}\textbf{67.702} 
& \cellcolor{top1color}\textbf{0.076}  & \cellcolor{top1color}\textbf{0.028}  & \cellcolor{top2color}\uline{0.829}   & \cellcolor{top2color}\uline{52.315} 
& \cellcolor{top1color}\textbf{0.051}  & \cellcolor{top3color}0.056  & \cellcolor{top1color}\textbf{0.886}   & \cellcolor{top2color}\uline{70.762}
& \cellcolor{top1color}\textbf{0.080}  & \cellcolor{top1color}\textbf{0.028}  & \cellcolor{top3color}0.704   & \cellcolor{top2color}\uline{34.517}  \\
\bottomrule
\end{tabular}}
\caption{Experimental results derived from~\texttt{Baichuan2-13B-Chat}.}
\label{table: baichuan2-13b-chat}
\vspace{-0.5em}
\end{table*}

\subsection{Confidence Estimation}
Given the model's Uncertainty for a given question $\mathcal{Q}$ and the fidelity $\mathbf{F}(\cdot)$ among different candidate answers, 
the confidence of the model in its answer $a_i$ for question $\mathcal{Q}$ is defined as follows: 
\begin{align}
    \mathbf{Conf}(\mathcal{Q}, a_i) &= \big( 1 - \mathbf{Uncertainty}(\mathcal{Q})\big)\cdot \mathbf{F}(a_i), \label{ali: conf}
\end{align}

\section{Experiments}
To validate the effectiveness of our proposed method, we conducted experiments on different RLHF-LMs such as \texttt{GPT-3.5-Turbo}\footnote{\href{https://openai.com/chatgpt}{https://openai.com/chatgpt}}, \texttt{GPT-4-Turbo}~\cite{openai2023gpt4}, \texttt{LLaMA2-Chat}~\cite{touvron2023llama} and \texttt{Baichuan2-13B-Chat}~\cite{yang2023baichuan}. 
To mitigate the influence of the sampling algorithm, unless specifically stated otherwise, we use hyper-parameters with a temperature of 1.0 and set top\_p as 1.0.
\subsection{Experimental Setting}
\paragraph{Dataset.} We have conducted experiments on four MCQA datasets to verify the effectiveness of our proposed confidence estimation method. 
ARC~\cite{Clark2018ThinkYH} is a dataset of 7,787 grade-school-level questions. We use the test split of the ARC-Challenge with 1,172 questions for our experiments.
MMLU~\cite{hendrycks2021measuring} is a dataset designed to measure knowledge acquired during pretraining and covers 57 subjects. To reduce the cost of API calls, we sampled $\frac{1}{8}$ of the data for testing for each subject. 
CommonSenseQA~\cite{talmor-etal-2019-commonsenseqa} is a dataset for commonsense question answering, and we use the validation split with 1,221 questions for experiments. 
TruthfulQA~\cite{lin-etal-2022-truthfulqa} is a dataset that contains 817 questions designed to evaluate language models’ preference to mimic some human falsehoods.
All the experiments are conducted under a $0$-shot setting.
\begin{table*}[!ht]
\centering
\small
\resizebox{\textwidth}{!}{\begin{tabular}{lccccccccccccccccc}
\toprule
& \multicolumn{4}{c}{\textbf{ARC-Challenge}} &  \multicolumn{4}{c}{\textbf{MMLU}} & \multicolumn{4}{c}{\textbf{CommonSenseQA}} &  \multicolumn{4}{c}{\textbf{TruthfulQA}} \\
\cmidrule(lr){2-5} \cmidrule(lr){6-9} \cmidrule(lr){10-13} \cmidrule(lr){14-17}
\textbf{Method} 
& \makecell{$\mathrm{ECE}_{10}$ \scalebox{0.65}{$\downarrow$}} & \makecell{$\mathrm{IPR}_{10}$ \scalebox{0.65}{$\downarrow$}} & \makecell{$\mathrm{CE}_{10}$ \scalebox{0.65}{$\uparrow$}} & \makecell{\textbf{Acc} \scalebox{0.65}{$\uparrow$}}  & 
 \makecell{$\mathrm{ECE}_{10}$ \scalebox{0.65}{$\downarrow$}} & \makecell{$\mathrm{IPR}_{10}$ \scalebox{0.65}{$\downarrow$}} & \makecell{$\mathrm{CE}_{10}$ \scalebox{0.65}{$\uparrow$}} & \makecell{\textbf{Acc} \scalebox{0.65}{$\uparrow$}}  & 
 \makecell{$\mathrm{ECE}_{10}$ \scalebox{0.65}{$\downarrow$}} & \makecell{$\mathrm{IPR}_{10}$ \scalebox{0.65}{$\downarrow$}}& \makecell{$\mathrm{CE}_{10}$ \scalebox{0.65}{$\uparrow$}}  & \makecell{\textbf{Acc} \scalebox{0.65}{$\uparrow$}} &  \makecell{$\mathrm{ECE}_{10}$ \scalebox{0.65}{$\downarrow$}} & \makecell{$\mathrm{IPR}_{10}$ \scalebox{0.65}{$\downarrow$}} & \makecell{$\mathrm{CE}_{10}$ \scalebox{0.65}{$\uparrow$}} & \makecell{\textbf{Acc} \scalebox{0.65}{$\uparrow$}} &  \\ 
\midrule
\multicolumn{17}{c}{\textsc{LLaMA2-7B-Chat}} \\ 
\midrule
Verb
& \cellcolor{top3color}0.294  & \cellcolor{top1color}\textbf{0.083}  & \cellcolor{top3color}0.482   & \cellcolor{top4color}45.904  
& \cellcolor{top4color}0.325  & \cellcolor{top3color}0.267  & \cellcolor{top4color}0.531   & \cellcolor{top4color}41.551 
& \cellcolor{top3color}0.208  & \cellcolor{top5color}0.267  & \cellcolor{top3color}0.516   & \cellcolor{top4color}52.662 
& \cellcolor{top4color}0.499  & \cellcolor{top3color}0.200  & \cellcolor{top3color}0.626   & \cellcolor{top5color}21.787  \\
Ling
& \cellcolor{top5color}0.452  & \cellcolor{top5color}0.333  & \cellcolor{top5color}0.283   & \cellcolor{top5color}44.625  
& \cellcolor{top5color}0.478  & \cellcolor{top5color}0.357  & \cellcolor{top5color}0.315   & \cellcolor{top5color}38.542  
& \cellcolor{top5color}0.385  & \cellcolor{top4color}0.250  & \cellcolor{top5color}0.275   & \cellcolor{top5color}51.597 
& \cellcolor{top5color}0.647  & \cellcolor{top5color}0.607  & \cellcolor{top5color}0.406   & \cellcolor{top4color}24.113  \\
Sampled
& \cellcolor{top4color}0.329  & \cellcolor{top3color}0.156  & \cellcolor{top2color}\uline{0.781}   & \cellcolor{top1color}\textbf{50.683}  
& \cellcolor{top3color}0.316  & \cellcolor{top2color}\uline{0.222}  & \cellcolor{top1color}\textbf{0.900}   & \cellcolor{top1color}\textbf{43.056}  
& \cellcolor{top4color}0.294  & \cellcolor{top3color}0.178  & \cellcolor{top2color}\uline{0.765}   & \cellcolor{top2color}\uline{54.627}  
& \cellcolor{top3color}0.389  & \cellcolor{top2color}\uline{0.133}  & \cellcolor{top1color}\textbf{0.875}   & \cellcolor{top1color}\textbf{27.540}  \\
Token
& \cellcolor{top2color}\uline{0.161}  & \cellcolor{top3color}0.156  & \cellcolor{top4color}0.430   & \cellcolor{top2color}50.256  
& \cellcolor{top2color}\uline{0.224}  & \cellcolor{top4color}0.333  & \cellcolor{top2color}\uline{0.593}   & \cellcolor{top2color}\uline{42.419}  
& \cellcolor{top2color}\uline{0.148}  & \cellcolor{top1color}\textbf{0.133}  & \cellcolor{top4color}0.417   & \cellcolor{top1color}\textbf{54.791}  
& \cellcolor{top2color}\uline{0.234}  & \cellcolor{top4color}0.289  & \cellcolor{top4color}0.484   & \cellcolor{top2color}\uline{27.417}  \\
\midrule
Ours
& \cellcolor{top1color}\textbf{0.073}  & \cellcolor{top2color}\uline{0.111}  & \cellcolor{top1color}\textbf{0.921}   & \cellcolor{top1color}\textbf{50.683}   
& \cellcolor{top1color}\textbf{0.102}  & \cellcolor{top1color}\textbf{0.167}  & \cellcolor{top2color}\uline{0.890}   & \cellcolor{top1color}\textbf{43.056}   
& \cellcolor{top1color}\textbf{0.053}  & \cellcolor{top2color}\uline{0.167}  & \cellcolor{top1color}\textbf{0.903}   & \cellcolor{top2color}\uline{54.627}    
& \cellcolor{top1color}\textbf{0.121}  & \cellcolor{top1color}\textbf{0.083}  & \cellcolor{top2color}\uline{0.762}   & \cellcolor{top1color}\textbf{27.540}  \\
\midrule
\multicolumn{17}{c}{\textsc{LLaMA2-13B-Chat}} \\ 
\midrule
Verb
& \cellcolor{top3color}0.198 & \cellcolor{top2color}\uline{0.143} & \cellcolor{top3color}0.495   & \cellcolor{top4color}57.594
& \cellcolor{top3color}0.286 & \cellcolor{top2color}\uline{0.214} & \cellcolor{top3color}0.572   & \cellcolor{top4color}45.614
& \cellcolor{top3color}0.204 & \cellcolor{top5color}0.278 & \cellcolor{top3color}0.497   & \cellcolor{top5color}56.260
& \cellcolor{top3color}0.443 & \cellcolor{top2color}\uline{0.167} & \cellcolor{top3color}0.732   & \cellcolor{top4color}27.138  \\
Ling
& \cellcolor{top5color}0.327 & \cellcolor{top5color}0.333 & \cellcolor{top5color}0.393   & \cellcolor{top5color}57.301 
& \cellcolor{top5color}0.448 & \cellcolor{top5color}0.333 & \cellcolor{top5color}0.378   & \cellcolor{top5color}45.040 
& \cellcolor{top5color}0.316 & \cellcolor{top2color}\uline{0.133} & \cellcolor{top4color}0.449   & \cellcolor{top4color}56.692
& \cellcolor{top5color}0.627 & \cellcolor{top5color}0.733 & \cellcolor{top4color}0.508   & \cellcolor{top5color}26.864 \\
Sampled
& \cellcolor{top4color}0.297 & \cellcolor{top4color}0.200 & \cellcolor{top2color}\uline{0.653}   & \cellcolor{top1color}\textbf{60.239}   
& \cellcolor{top4color}0.351 & \cellcolor{top4color}0.267 & \cellcolor{top2color}\uline{0.788}   & \cellcolor{top2color}\uline{47.251}  
& \cellcolor{top4color}0.287 & \cellcolor{top3color}0.156 & \cellcolor{top2color}\uline{0.717}   & \cellcolor{top1color}\textbf{58.722}  
& \cellcolor{top4color}0.461 & \cellcolor{top4color}0.422 & \cellcolor{top1color}\textbf{0.798}   & \cellcolor{top2color}\uline{29.131} \\
Token
& \cellcolor{top2color}\uline{0.135} & \cellcolor{top3color}0.178 & \cellcolor{top4color}0.408   & \cellcolor{top2color}\uline{59.898} 
& \cellcolor{top2color}\uline{0.225} & \cellcolor{top3color}0.244 & \cellcolor{top4color}0.502   & \cellcolor{top1color}\textbf{47.512} 
& \cellcolor{top2color}\uline{0.142} & \cellcolor{top4color}0.222 & \cellcolor{top5color}0.403   & \cellcolor{top2color}\uline{57.007} 
& \cellcolor{top2color}\uline{0.238} & \cellcolor{top3color}0.200 & \cellcolor{top5color}0.429   & \cellcolor{top1color}\textbf{30.845} \\
\midrule
Ours
& \cellcolor{top1color}\textbf{0.069} & \cellcolor{top1color}\textbf{0.111} & \cellcolor{top1color}\textbf{0.886}   & \cellcolor{top1color}\textbf{60.239}  
& \cellcolor{top1color}\textbf{0.070} & \cellcolor{top1color}\textbf{0.083} & \cellcolor{top1color}\textbf{0.852}   & \cellcolor{top2color}\uline{47.251}  
& \cellcolor{top1color}\textbf{0.043} & \cellcolor{top1color}\textbf{0.083} & \cellcolor{top1color}\textbf{0.883}   & \cellcolor{top1color}\textbf{58.722}     
& \cellcolor{top1color}\textbf{0.121} & \cellcolor{top1color}\textbf{0.083} & \cellcolor{top2color}\uline{0.762}   & \cellcolor{top2color}\uline{29.131}  \\
\midrule
\multicolumn{17}{c}{\textsc{LLaMA2-70B-Chat}} \\ 
\midrule
Verb
& \cellcolor{top1color}\textbf{0.071}  & \cellcolor{top3color}0.286  & \cellcolor{top3color}0.369   & \cellcolor{top4color}70.819 
& \cellcolor{top3color}0.236  & \cellcolor{top2color}\uline{0.194}  & \cellcolor{top3color}0.351   & \cellcolor{top4color}53.183 
& \cellcolor{top1color}\textbf{0.069}  & \cellcolor{top5color}0.222  & \cellcolor{top4color}0.286   & \cellcolor{top4color}70.680 
& \cellcolor{top3color}0.311  & \cellcolor{top1color}\textbf{0.028}  & \cellcolor{top3color}0.522   & \cellcolor{top4color}43.452   \\
Ling
& \cellcolor{top5color}0.223  & \cellcolor{top5color}0.333  & \cellcolor{top5color}0.119   & \cellcolor{top5color}67.833 
& \cellcolor{top5color}0.375  & \cellcolor{top4color}0.333  & \cellcolor{top5color}0.096   & \cellcolor{top5color}51.794  
& \cellcolor{top4color}0.189  & \cellcolor{top1color}\textbf{0.067}  & \cellcolor{top5color}0.117   & \cellcolor{top5color}70.106 
& \cellcolor{top5color}0.507  & \cellcolor{top5color}0.400  & \cellcolor{top5color}0.289   & \cellcolor{top5color}36.597  \\
Sampled
& \cellcolor{top4color}0.220  & \cellcolor{top4color}0.311  & \cellcolor{top2color}\uline{0.475}   & \cellcolor{top2color}\uline{72.867}  
& \cellcolor{top4color}0.325  & \cellcolor{top3color}0.289  & \cellcolor{top4color}0.289   & \cellcolor{top2color}\uline{56.308}  
& \cellcolor{top5color}0.212  & \cellcolor{top2color}\uline{0.089}  & \cellcolor{top2color}\uline{0.551}   & \cellcolor{top1color}\textbf{72.809} 
& \cellcolor{top4color}0.351  & \cellcolor{top3color}0.156  & \cellcolor{top2color}\uline{0.622}   & \cellcolor{top2color}\uline{51.897}  \\
Token
& \cellcolor{top3color}0.091  & \cellcolor{top2color}\uline{0.200}  & \cellcolor{top4color}0.315   & \cellcolor{top1color}\textbf{73.208} 
& \cellcolor{top2color}\uline{0.190}  & \cellcolor{top5color}0.378  & \cellcolor{top2color}\uline{0.378}   & \cellcolor{top1color}\textbf{56.597}  
& \cellcolor{top2color}\uline{0.093}  & \cellcolor{top4color}0.178  & \cellcolor{top3color}0.339   & \cellcolor{top2color}\uline{72.645} 
& \cellcolor{top2color}\uline{0.173}  & \cellcolor{top4color}0.267  & \cellcolor{top4color}0.352   & \cellcolor{top1color}\textbf{52.020}  \\
\midrule
Ours
& \cellcolor{top2color}\uline{0.085}  & \cellcolor{top1color}\textbf{0.111}  & \cellcolor{top1color}\textbf{0.908}   & \cellcolor{top2color}\uline{72.867} 
& \cellcolor{top1color}\textbf{0.066}  & \cellcolor{top1color}\textbf{0.083}  & \cellcolor{top1color}\textbf{0.898}   & \cellcolor{top2color}\uline{56.308}  
& \cellcolor{top3color}0.094  & \cellcolor{top3color}0.111  & \cellcolor{top1color}\textbf{0.918}   & \cellcolor{top1color}\textbf{72.809} 
& \cellcolor{top1color}\textbf{0.093}  & \cellcolor{top2color}\uline{0.089}  & \cellcolor{top1color}\textbf{0.804}   & \cellcolor{top2color}\uline{51.897}  \\

\bottomrule
\end{tabular}}
\caption{Experimental results derived from \texttt{LLaMA-2-Chat}.}
\vspace{-3mm}
\label{table: llama2-13b-chat}
\end{table*}

\paragraph{Metrics.} 
We utilize multiple metrics to evaluate. 
We bin the predictions from the model by their confidence and report the $\mathrm{ECE}$~(expected calibration error). We also report the Brier Score of different methods in Table~\ref{table: brierscore}.
In this paper, we also defines two novel metrics to evaluate the calibration.
The first one is $\mathrm{IPR}$ (\textbf{I}nverse \textbf{P}air \textbf{R}atio), which is used to measure the monotonicity of the reliability diagram.
If the reliability diagram is monotonic, it indicates that the average accuracy of answers with low confidence is lower than the average accuracy of answers with high confidence.
\begin{align}
     \mathrm{IPR}_{M} = \frac{\mathrm{IP}}{C_{K}^2},
\end{align}
where $\mathrm{IP}$ is the inverse pair number in the reliable diagram, and $K$ is the bin number with a density larger than 0.
We found that as the number of model parameters increases, the accuracy of the model improves across various datasets. However, language models still tend to consistently express uncertainty within certain fixed ranges, and ECE cannot clearly reflect this phenomenon.
Therefore, we suggest using the $\mathrm{CE}$~(\textbf{C}onfidence \textbf{E}venness) to evaluate the uniformity of the density of each bar in the reliability diagram.

\begin{align}
    \mathrm{CE}_{M} = - \frac{\sum^{\mathrm{M}}_{i=1}p_i\cdot\log p_i}{\log\mathrm{M}},
\end{align}
In this paper, we adopt $10$ equal-size bins to calculate $\mathrm{ECE}_{10}$, $\mathrm{IPR}_{10}$ and $\mathrm{CE}_{10}$. 
We also report the accuracy on these benchmarks to measure whether calibration reduces the accuracy.

\paragraph{Baselines.} 
We compared our approach with different baselines for eliciting the confidence of language model. 
First, we reproduced the \textbf{Verb} and \textbf{Ling} method proposed by~\citet{tian-etal-2023-just}.
The \textbf{Verb} method involves prompting the model to output a floating-point number between 0 and 1 to represent its confidence immediately after providing an answer~\cite{tian-etal-2023-just, lin2022teaching}. 
The \textbf{Ling} method entails having the language model express its confidence level in natural language~\cite{tian-etal-2023-just}. 
Since commercial models like \texttt{ChatGPT} do not provide per-token logits, we employed a sampling technique to estimate the probability of token generation, referred to as the \textbf{Sampled} method. 
Unless otherwise specified, the Sampled method involves sampling 10 times. 
For open-source models like \texttt{LLaMA2-Chat}, we directly use the probability of token generation as the measure of the language model's confidence, which we refer to as the \textbf{Token} method.
We also compare the \textbf{Conformal} Prediction Baseline proposed by~\citet{kumar2023conformal} with our UF calibration in Appendix~\ref{sec: comparewithconformal}.
All the prompt templates we use are shown in Appendix~\ref{sec: prompt}.

\subsection{Main Results} \label{main_results}
Tables~\ref{table: chatgpt}--\ref{table: llama2-13b-chat} show our experimental results on \texttt{GPT-3.5-Turbo}, \texttt{GPT-4-Turbo}, \texttt{Baichuan2-13B-Chat}, and \texttt{LLaMA2-Chat}. 
Based on the experimental results, the following conclusions can be drawn:
\vspace{-0.5em}
\begin{enumerate}[label={\arabic*)}, leftmargin=*]
\item Our proposed method demonstrates a clear improvement over the various baselines in terms of three metrics: $\mathrm{ECE}_{10}$, $\mathrm{IPR}_{10}$, and $\mathrm{CE}_{10}$, which demonstrates the effectiveness of our method.
\vspace{-0.5em}
\item The Verb and Ling methods might, to some extent, impair the language model's accuracy on multiple-choice question answering tasks, which might be caused by more complicated instructions.
Additionally, since the Ling method is more complex, it has a greater impact on the overall accuracy than the Verb method.
\vspace{-0.5em}
\item Similar to the conclusion from~\citet{tian-etal-2023-just}, the calibration of the Verb method tends to be better than that of the Ling method. This is because the linguistic expressions used in the Ling method are based on human psychology. However, the confidence represented by the same expression may have a gap between humans and models and among different models and different
sentences might mean the same thing~\cite{kuhn2023semantic}. 
\vspace{-0.5em}
\item The $\mathrm{CE}_{10}$ of the Verbalization-based Method is relatively low, which suggests that language models tends to prefer outputting expressions of certain confidence, such as \texttt{`Highly Likely'}, 0.8 and 0.9. 
This phenomenon can also explain why the $\mathrm{ECE}_{10}$ of the Verbalization-based Method improves when the overall average accuracy of the model is between 70-90\%. 
\end{enumerate}
\vspace{-0.3cm}

\begin{figure*}[!t]
  \centering
  \includegraphics[width=0.9\textwidth]{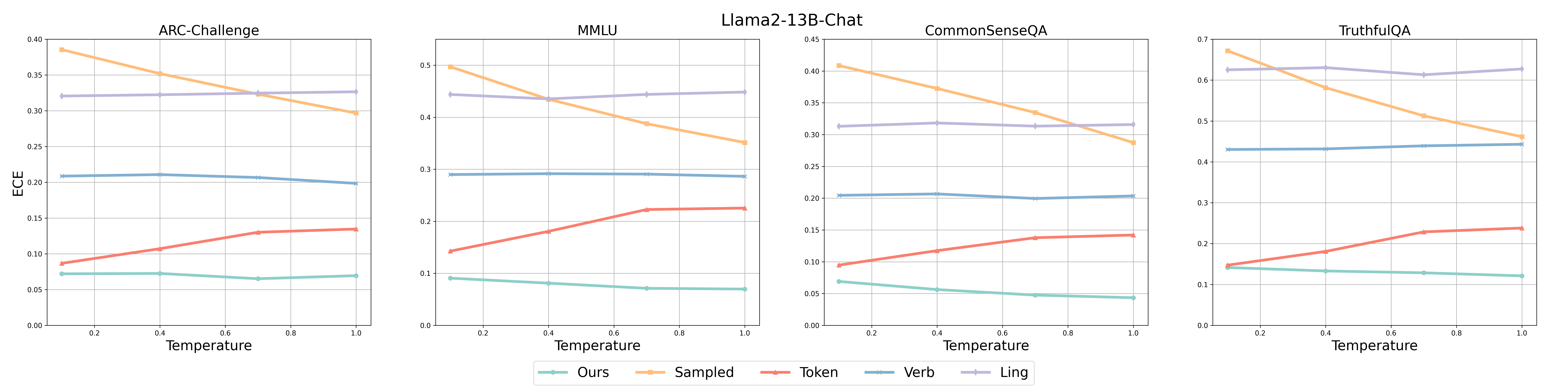}
  \vspace{-0.5em}
  \caption{Our proposed method achieved well-calibrated results across all temperatures. The experimental results are derived from \texttt{LLaMA2-13B-Chat}. The results from \texttt{Baichuan2-13B-Chat} are presented in Appendix Figure~\ref{fig:baichuan_temperature_scaling}.} \label{fig:temperature_scaling}
\end{figure*}

\begin{figure*}[!t]
  \centering
  \includegraphics[width=0.9\textwidth]{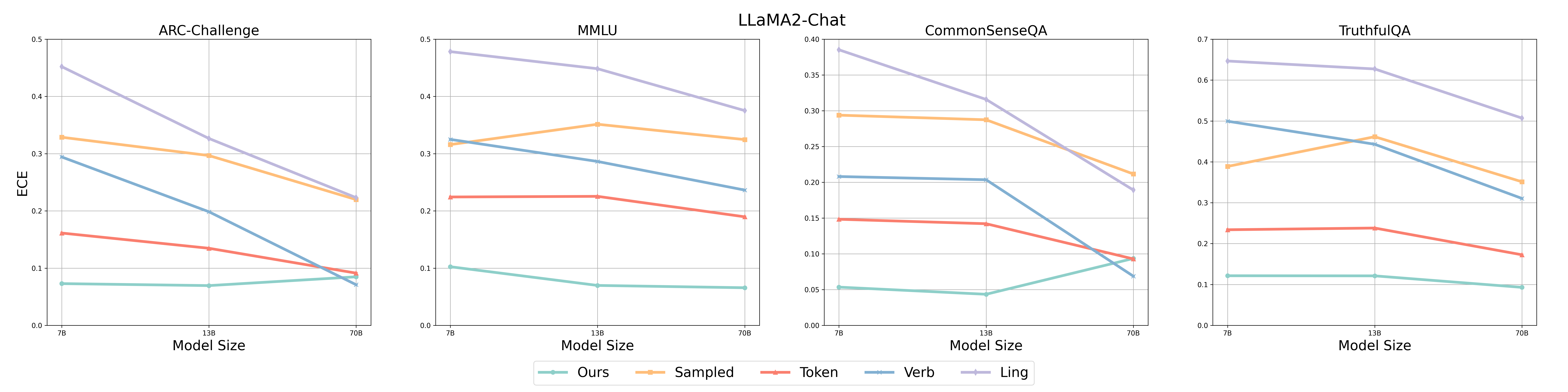}
  \vspace{-0.5em}
  \caption{The experimental results are derived from  \texttt{LLaMA2-Chat}.} \label{fig: parameter_scaling}
\end{figure*}

\subsection{Ablation Study}
As shown in Table~\ref{tab:Ablation study}, removing $\mathrm{Uncertainty}$ and only relying on $\mathrm{Fidelity}$ to estimate the model's confidence, we can also achieve comparatively better calibration than other methods.
This phenomenon indicates that our proposed method reflects the language model's Fidelity to its answers very well.
Meanwhile, it is difficult to estimate the model's confidence only depending on $\mathrm{Uncertainty}$. As mentioned in~\ref{sec:Uncertainty_Estimation}, $\mathrm{Uncertainty}$ is designed for measuring the model's uncertainty regarding the question $\mathcal{Q}$, rather than its confidence for a particular answer.
In the section~\ref{sec:method_fidelity}, we utilize (\ref{ali: fidelity}) to calculate the language model's normalized fidelity in a hierarchical fidelity chain, where $\tau$ is a hyper-parameter.
The larger the value of $\tau$, the lower the estimated fidelity for answers closer to the end of the fidelity chain.
Our experiments in Table~\ref{tab:Ablation study} indicate that setting $\tau$ to around 2 is a relatively appropriate choice for the fidelity estimation process.
If $\tau$ is too large, the $\mathrm{ECE}_{10}$ will also increase, which will cause the issue of overconfidence of our estimated confidence.

\begin{table}[h]
\centering
\resizebox{1.0\columnwidth}{!}{
\begin{small}
\begin{tabular}{lccccc}
\toprule
\bf Method & \bf ARC & \bf MMLU & \bf CSQA & \bf TruthfulQA& \bf Avg.\\
\midrule
\textbf{Ours} & \bf 0.069 & \bf 0.070 & \bf 0.043 & \bf 0.121 & \bf 0.076\\
\cmidrule(lr){1-6}
w/o. Uncertainty        & 0.122 & 0.184 & 0.115 & 0.202  & 0.156\\
w/o. Fidelity           & 0.675 & 0.614 & 0.704 & 0.677  & 0.668\\
\cmidrule(lr){1-6}
$\tau = 1.5$            & 0.103 & \bf 0.064 & 0.066 & \bf 0.082  & 0.079\\
\rowcolor{gray!20}
$\tau = 2.0$ (Default)  & 0.069 & 0.070 & 0.043 & 0.121 & \bf 0.076\\
$\tau = 2.5$            & \bf 0.067 & 0.089 & \bf 0.040 & 0.142  & 0.085\\
$\tau = 3.0$            & 0.074 & 0.107 & 0.050 & 0.155  & 0.097\\
$\tau = 4.0$            & 0.085 & 0.138 & 0.075 & 0.165  & 0.116\\
$\tau = 5.0$            & 0.102 & 0.158 & 0.094 & 0.183  & 0.134\\
\cmidrule(lr){1-6}
Best Result (Others) & 0.135 & 0.225 & 0.142 & 0.238  & 0.185\\
\bottomrule
\end{tabular}
\end{small}
 }
\caption{Ablation study of our method. The results ($\mathrm{ECE}_{10}$) are derived from \texttt{LLaMA2-13B-Chat}.}
\label{tab:Ablation study}
\vspace{-10pt}
\end{table}

\section{Analysis and Discussion}
To take a closer look at the difference between different calibration methods tailored for language models, in this section, we verify the robustness of our method from two aspects: \textit{Temperature-Scaling} and \textit{Parameter-Scaling}. Meanwhile, we also conducted a detailed discussion of a research question: \textit{What kind of Confidence is Truly Well-Calibrated?}

\begin{figure*}[!t]
  \centering
  \includegraphics[width=0.9\textwidth]{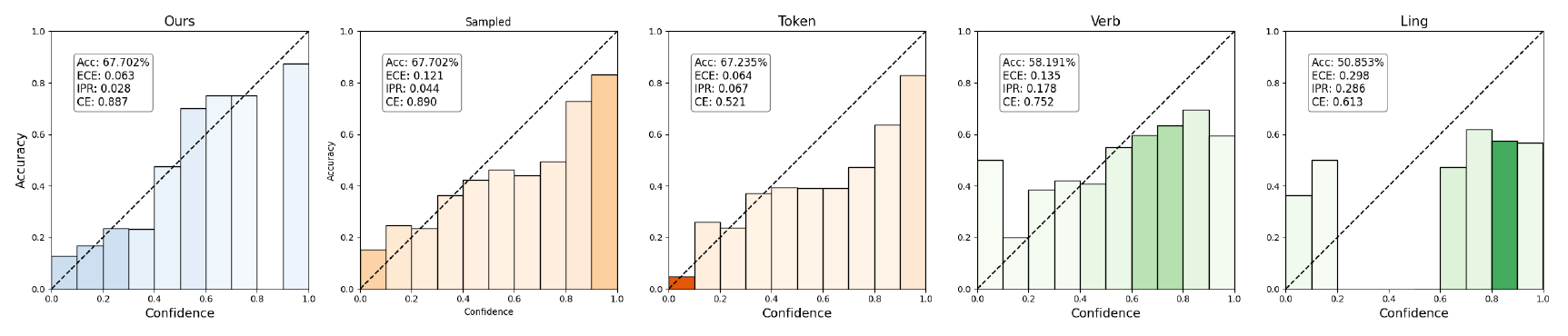}
  \vspace{-0.5em}
  \caption{Reliability diagrams of \texttt{Baichuan2-13B-Chat} on ARC-Challenge. In these diagrams, the darker the color, the higher the density. The reliability diagrams of other models we evaluated are shown in Appendix Figures~\ref{fig:full-gpt_3.5_turbo}--\ref{fig:full-llama2_70b_chat}.} \label{fig:well-calibrated}
\vspace{-10pt}
\end{figure*}

\paragraph{Temperature-Scaling}
In the main experiments, we evaluate various methods using a constant temperature of 1.0.
In this section, we will explore the influence of sampling temperature on the performance of different methods.
As illustrated in Figures~\ref{fig:temperature_scaling} and~\ref{fig:baichuan_temperature_scaling}, our proposed calibration method consistently achieves the lowest expected calibration error across all temperatures, showing remarkable robustness to temperature variations. 
This is because, in eliciting model fidelity, our method always employs Greedy Decoding rather than Sampling. 
Thus, the hierarchical chains we obtain are usually consistent across different sampling temperatures. 
In contrast, the expected calibration error of Logit-based Methods is usually affected by temperature. 
For the Sampling method with limited sampling budgets, the lower the temperature, the more significantly the diversity of the sampled results will decrease, exacerbating the overconfidence of language models. 
For the Token Method, the impact of temperature on its calibration shows a trend of ``\textit{first increasing and then remaining relatively stable}'' or ''\textit{first increasing and then decreasing}``.
This is because we could directly utilize (\ref{ali: logit-based method}) to estimate the confidence of each option, and if the temperature is too low (i.e., 0.1), it will lead to the confidence of a large number of options approaching zero.
This phenomenon might contribute to reducing expected calibration error, but it does not necessarily indicate that the model's confidence is well-calibrated.
The Verbalization-based method is less affected by temperature, which indicates that the expressions which language models prefer to output are relatively consistent across different temperatures.
\vspace{-0.5em}
\paragraph{Parameter-Scaling}
As shown in Figure~\ref{fig: parameter_scaling}, we evaluate the calibration of various methods at different parameter scales on the \texttt{LLaMA2-Chat} series models.
Our proposed method exhibits good calibration across different amounts of model parameters.
With the size of model parameters increasing, the calibration of the Verbalization-based method and the Logit-based method is improving.
This phenomenon indicates that as the scale of model parameters increases, the model's Self-Awareness is improving.
However, the relatively high expected calibration error suggests that language models still have issues with overconfidence.
\vspace{-0.5em}
\paragraph{Truly Well-Calibrated Confidence}
Previous work mainly evaluates the calibration of language models through ECE. 
This section will discuss the research question: ``\textit{What Kind of Confidence is Truly Well-Calibrated?}''.
Figure~\ref{fig:well-calibrated} demonstrates the calibration of various methods.
From the calibration perspective, we hope that the confidence and accuracy relationship is close to the curve $y = x$. Thus, we need to reduce the ECE by calibrating confidence.
Meanwhile, we hope that the reliability diagram should be as monotonic as possible to ensure that the accuracy of the results generated with low confidence is lower than that of the results with high confidence. 
Therefore, we propose the \textit{Inverse Pair Ratio}~(IPR) to evaluate monotonicity.
From the perspective of building a more honest system, we hope the model's confidence should be distributed across different confidence intervals.
For example, if a language model has an overall accuracy of 75\% on the TruthfulQA dataset and the confidence of each question from the language model is always 75\%, its ECE and IPR would be 0.
And we find that different models tend to express confidence within a fixed interval.
In this case, we think that the confidence may not necessarily be a truly well-calibrated confidence because we could not exclude some low-confidence results based on the confidence from the language model.
Although the prior distribution of the model’s confidence is unknown, our confidence estimation method finds that language models have different confidence for different questions.
Thus, we propose a metric called~\textit{Confidence Evenness}~(CE) to measure whether the model confidence always is located in a fixed interval.
We believe ECE, IPR, and CE evaluate calibration from different perspectives and there is a trade-off between these three metrics.
We suggest that truly well-calibrated confidence should achieve a balance among ECE, IPR, and CE, rather than over-optimizing any of them.

\section{Conclusion}
In this paper, we decompose the language model confidence into the \textit{Uncertainty} about the question and the \textit{Fidelity} to the answer generated by language models. 
Through the decomposition, we propose a plug-and-play method, \textsc{UF Calibration}, to calibrate the confidence of language models.
Through experiments with 6 RLHF-LMs on 4 multiple-choice question answering benchmarks, our method exhibits good calibration.
Besides, we propose two novel metrics, $\mathrm{IPR}$ and $\mathrm{CE}$, to evaluate the calibration of language models.
Finally, we conduct a detailed discussion on \textit{Truly Well-Calibrated Confidence}.
We believe our method can serve as a strong baseline, and we hope that this work could provide some insights into the language model confidence calibration.

\section*{Limitations}
Although our method has shown good calibration, it is mainly applicable to scenarios where the set of answers is known, i.e., multiple-choice question answering, text classification, sentiment classification, and preference labeling in RLHF.
Eliciting the model’s fidelity in open-ended generation scenarios is a direction worth exploring.
Meanwhile, our method involves multiple invocations of language models, and how to estimate the probability distribution of tokens generated by the language model with as few callings as possible remains to be studied.

\section*{Acknowledgements}
This work was supported by the National Natural Science Foundation of China (No. 62441602).
The computations in this research were performed using the CFFF platform of Fudan University.

\bibliography{custom}

\appendix
\section{The Computation Cost of Eliciting Fidelity} 
In this section, we will display the average length of the fidelity chains for different models across various datasets in the Table~\ref{table: averageLength}.
Since we deploy greedy decoding during the process of eliciting fidelity,the average length of the fidelity chain is equal to the average number of requests. At the same time, it should be noted that, when eliciting the Fidelity Chain, only 1 token needs to be generated. 
Therefore, the average length of the fidelity chain can also be regarded as the average number of tokens generated.

\begin{table}[h]
\centering
\small
\resizebox{\columnwidth}{!}{\begin{tabular}{lccccc}
\toprule
\textbf{Model} & \textbf{ARC-Challenge} & \textbf{MMLU}  & \textbf{CommonSenseQA} & \textbf{TruthfulQA} & \textbf{Avg.} \\ 
\midrule
\textsc{GPT-3.5-Turbo} & 2.774& 2.984& 3.052& 3.275& 3.021\\
\textsc{GPT-4-Turbo} & 1.492& 1.915& 2.157& 1.616& 1.795 \\
\textsc{Baichuan2-13B-Chat}  & 2.830& 2.820& 2.889& 4.345& 3.221\\
\textsc{Llama2-7B-Chat}  & 2.467& 2.631& 2.771& 3.944& 2.953\\
\textsc{Llama2-13B-Chat} & 2.725& 2.875& 2.956& 4.100& 3.164\\
\textsc{Llama2-70B-Chat}  & 2.384& 2.563& 2.455& 3.284& 2.671 \\
\bottomrule
\end{tabular}}
\caption{The average length of the fidelity chains for different models across various datasets}
\label{table: averageLength}
\vspace{-0.5em}
\end{table}

\section{Algorithm}
The pseudo code of our proposed method is shown in Algorithm~\ref{alg: method}.
It should be clarified that, as long as a candidate answer $a_i$ appears in the answer set $\mathcal{A}$ or the Fidelity chain set $\mathcal{S}$, we could estimate its confidence through (\ref{ali: conf}).

\begin{algorithm}[!ht]
\caption{Algorithm}\label{alg: method}
\begin{algorithmic}[1]
\Require Input question $\mathcal{Q}$, Option list $\mathcal{O}$, Answer set $\mathcal{A} = \varnothing$, Sampling budget $K$, RLHF-LM~$\mathrm{LM}$, $o^*$ is ``All other options are wrong.'', Fidelity chain set $\mathcal{S}$, $\mathbf{U}(\cdot)$ refers to~(\ref{ali: uncertainty}).
\State $t \gets 0$
\While{$t < K$}
\State $a_i \gets \mathrm{LM}(Q, \mathcal{O})$
\Comment{Sampling answer}
\State $\mathcal{A} \gets \mathcal{A} \cup\{a_i\}$
\State $\mathcal{P}_{sampled}(a_i) \gets \mathcal{P}_{sampled}(a_i) + 1$ 
\State $t \gets t + 1$
\Comment{Continue sampling }
\EndWhile 
\State $\mathcal{P}_{sampled}(a_i) \gets \mathcal{P}_{sampled}(a_i) / K$ \\
\State $\mathbf{Uncertainty}(\mathcal{Q}) = \mathbf{U}(\mathcal{P}_{sampled})$ 

\Comment{Get uncertainty}
\State $i \gets 0$
\While{$ |\mathcal{A} | > 0$}
\State $\mathcal{A} \gets A\setminus \{a_i\}$
\Comment{Select a answer}
\State $\mathcal{O}^* \gets (\mathcal{O} \setminus \{o_i\}) \cup o_*$
\Comment{Replace option}
\State $\mathcal{C}_i = a_i$
\Comment{Init a fidelity chain}
\While{$|\mathcal{O}^*| > 0$}
\State $a^* \gets \mathrm{LM}(\mathcal{Q}, \mathcal{O}^*)$
\Comment{Greedy decoding}
\If{$a^* \neq a_i$}    
\Comment{Low fidelity}
    \State $\mathcal{O}^* \gets \mathcal{O}^* \setminus \{o_i\}$
    \Comment{Delete option}
    \State $a_i = a^*$
    \State $\mathcal{C}_i = (\mathcal{C}_i \rightarrow a_*)$
    \Comment{Add element}
\Else
    \State break 
    \Comment{High fidelity}
\EndIf
\EndWhile 
\State $\mathcal{S} \gets \mathcal{S} \cup \mathcal{C}_i$
\State $i \gets i + 1$
\EndWhile \\
\State $\mathbf{F}(a_i) = \sum_{j=1}^{|\mathcal{A}|} \mathcal{P}_{\mathrm{sampled}}({\mathcal{C}_j})\cdot \mathbf{Fidelity}_{\mathcal{C}_j}(a_i)$

\Comment{Get fidelity}
\State $\mathrm{Conf}(\mathcal{Q}, a_i) = (1 - \mathbf{Uncertainty}(\mathcal{Q})) \cdot \mathbf{F}(a_i)$
\Comment{Get confidence}\\
\Return $\mathrm{Conf}(\mathcal{Q}, a_i)$

\Comment{Return the confidence of answer $a_i$}
\end{algorithmic}
\end{algorithm}

\begin{table*}[!t]
\centering
\small
\resizebox{\textwidth}{!}{\begin{tabular}{lccc}
\toprule
\textbf{Model} & \textbf{Is the answer chosen in the first round correct?} & \textbf{Choose "All other options are wrong." after replacing} & \textbf{Do not choose "All other options are wrong." after replacing} \\ 
\midrule
	
\textsc{GPT-3.5-Turbo} & \textbf{True} & 25.99\% & 33.27\%\\
                       & \textbf{False} & 5.85\% & 34.88\%\\
\cmidrule(lr){2-4}
                      & \textbf{Acc.} & 81.61\% & 48.82\%\\
\cmidrule(lr){1-4}
\textsc{GPT-4-Turbo} & \textbf{True} & 70.75\%& 16.83\%\\
                       & \textbf{False} & 3.00\%& 9.42\%\\
\cmidrule(lr){2-4}
                      & \textbf{Acc.} & 95.93\%& 64.10\%\\
\cmidrule(lr){1-4}
\textsc{Baicuan2-13B-Chat} & \textbf{True} & 5.14\%& 29.40\%\\
                       & \textbf{False} & 4.22\%& 61.24\%\\
\cmidrule(lr){2-4}
                      & \textbf{Acc.} & 54.90\%& 32.43\%\\
\cmidrule(lr){1-4}	
\textsc{Llama2-7B-Chat} & \textbf{True} & 3.92\%& 23.50\%\\
                       & \textbf{False} & 4.83\%& 67.75\%\\
\cmidrule(lr){2-4}
                      & \textbf{Acc.} & 44.76\%& 25.75\%\\
\cmidrule(lr){1-4}
\textsc{Llama2-13B-Chat} & \textbf{True} & 3.55\%& 25.64\%\\
                       & \textbf{False} & 2.82\%& 67.99\%\\
\cmidrule(lr){2-4}
                      & \textbf{Acc.} & 55.77\%& 27.39\%\\
\cmidrule(lr){1-4}

\textsc{Llama2-70B-Chat} & \textbf{True} & 13.59\%& 38.43\%\\
                       & \textbf{False} & 3.98\%& 44.00\%\\
\cmidrule(lr){2-4}
                      & \textbf{Acc.} & 77.35\%& 46.62\%\\
\bottomrule
\end{tabular}}
\caption{We found that if the option chosen by the model in the first round is replaced with "All other options are wrong," the model then chooses "All other options are wrong" in the second round. In this case, the accuracy of the model's first-round choice is significantly higher compared to when it chooses other options in the second round. The results are derived from TruthfulQA.}
\label{table: whyfidelity}
\vspace{-0.5em}
\end{table*}

\begin{table*}[!t]
\centering
\small
\resizebox{\textwidth}{!}{\begin{tabular}{lcccccccccccccccccccc}
\toprule
\textbf{Dataset} & \textbf{Method} & \textbf{Model}  & $0.0$ & $0.02$ & $0.05$ & $0.1$ & $0.2$ & $0.25$ & $0.3$ & $0.4$ & $0.5$ & $0.6$ & $0.7$ & $0.8$ & $0.9$ & $0.95$ & $1.0$ & \makecell{$\mathrm{ECE}_{10}$ \scalebox{0.65}{$\downarrow$}} &  \makecell{$\mathrm{CE}_{10}$ \scalebox{0.65}{$\uparrow$}} & \makecell{\textbf{Acc} \scalebox{0.65}{$\uparrow$}}  \\ 
\midrule
CSQA & Verb & \textsc{LLaMA2-7B-Chat} &  3&    0&     0&    1&   25&    0&   23&   5&   78&   10&  309&  727&   19&   0&   21&  0.208&  0.516&   52.662\\
& & \textsc{LLaMA2-13B-Chat}  & 11&     0&     0&    0&    9&    0&    1&   29&    7&  112&  108&  851&   61&     0&   32&  0.204& 0.497&   56.260\\ 
& & \textsc{LLaMA2-70B-Chat} &    6&     0&     0&    2&    2&    0&    3&    3&    1&   23&  221&  955&    2&     0&    3&  0.069& 0.286&   70.680\\ 
\cmidrule(lr){2-3}  \cmidrule(lr){4-18}  \cmidrule(lr){19-21}
 & Ling & \textsc{LLaMA2-7B-Chat} &   11&     0&    21&    0&    3&    0&    0&    0&    1&    5&    2&   13& 1020&    75&   70&  0.385& 0.275&   51.597 \\
& & \textsc{LLaMA2-13B-Chat} &   18&     1&    11&    0&    6&    0&    0&    0&    0&    0&    3&  194&  892&    96&    0&  0.316& 0.449&   56.692 \\ 
& & \textsc{LLaMA2-70B-Chat} &    0&     0&    26&    0&    0&    0&    0&    0&    0&    1&    2&    2& 1172&     2&   16&  0.189& 0.117&   70.106 \\ 
\midrule
MMLU & Verb & \textsc{LLaMA2-7B-Chat}&   14&     0&     0&    3&   46&    0&   21&   16&   65&   44&  488&  981&   26&     0&   24&  0.325& 0.531&   41.551 \\
& & \textsc{LLaMA2-13B-Chat}&   23&     0&     0&    0&   41&    0&    0&   54&    7&  227&  278& 1056&   18&     0&   24&  0.286& 0.572&   45.614 \\ 
& & \textsc{LLaMA2-70B-Chat}&    1&     0&     0&    0&    7&    0&    3&    1&    2&    9&  518& 1159&    1&     0&   27&  0.236& 0.351&   53.183 \\ 
\cmidrule(lr){2-3}  \cmidrule(lr){4-18}  \cmidrule(lr){19-21}
 & Ling & \textsc{LLaMA2-7B-Chat}&   47&     0&   101&    0&   21&    0&    0&    0&    6&    4&    7&   12& 1408&    77&   45&  0.478& 0.315&   38.542 \\
& & \textsc{LLaMA2-13B-Chat}&   81&     1&    15&    0&    4&    2&    0&    0&    0&    0&    4&   84& 1261&   261&   11&  0.448& 0.378&   45.040 \\ 
& & \textsc{LLaMA2-70B-Chat}&    3&     0&    31&    0&    0&    0&    0&    0&    0&    6&    2&    5& 1673&     1&    7&  0.375& 0.096&   51.794 \\ 
\midrule
ARC & Verb & \textsc{LLaMA2-7B-Chat}&    4&     0&     0&    0&   26&    0&   13&    6&   53&    5&  216&  800&   20&     0&   29&  0.294& 0.482&   45.904 \\
& & \textsc{LLaMA2-13B-Chat}&    1&     0&     0&    0&   31&    0&    0&   13&   13&   68&  129&  851&   18&     0&   47&  0.198& 0.495&   57.594 \\ 
& & \textsc{LLaMA2-70B-Chat}&    3&     0&     0&    0&   11&    0&    3&    0&    2&    6&  288&  836&    3&     0&   20&  0.071& 0.369&   70.819 \\ 
\cmidrule(lr){2-3}  \cmidrule(lr){4-18}  \cmidrule(lr){19-21}
 & Ling & \textsc{LLaMA2-7B-Chat} &    3&     0&    24&    0&   10&    0&    0&    0&    0&    0&    5&   10& 1023&    53&   44&  0.452& 0.283&   44.625\\
& & \textsc{LLaMA2-13B-Chat}&    1&     0&     5&    0&    5&    0&    0&    0&    0&    0&    1&   76&  914&   162&    8&  0.327& 0.393&   57.301 \\ 
& & \textsc{LLaMA2-70B-Chat} &    3&     0&    27&    1&    0&    0&    0&    0&    0&    3&    1&    1& 1121&     2&   13&  0.223& 0.119&   67.833\\ 
\midrule
TruthfulQA & Verb & \textsc{LLaMA2-7B-Chat} &   10&     0&     0&    1&   23&    0&    8&    2&  125&   18&  167&  406&   17&     0&   40&  0.499& 0.626&   21.787 \\
& & \textsc{LLaMA2-13B-Chat}&   11&     0&     0&    1&   11&    0&    0&   56&   34&  145&  116&  369&   26&     0&   48&  0.443& 0.732&   27.138 \\ 
& & \textsc{LLaMA2-70B-Chat}&    3&     0&     0&    0&    7&    0&    4&    4&    4&   22&  320&  404&    9&     0&   30&  0.311& 0.522&   43.452 \\ 
\cmidrule(lr){2-3}  \cmidrule(lr){4-18}  \cmidrule(lr){19-21}
 & Ling & \textsc{LLaMA2-7B-Chat}&   30&     0&    53&    0&   10&    0&    0&    0&    8&    4&    4&   15&  611&    43&   39&  0.647& 0.406&   24.113 \\
& & \textsc{LLaMA2-13B-Chat}&   39&     2&    19&    0&    4&    0&    0&    0&    0&    0&    4&   40&  526&   177&    6&  0.627& 0.508&   26.864 \\ 
& & \textsc{LLaMA2-70B-Chat}&   10&     0&    31&    0&    0&    0&    0&    0&    0&    3&    0&    9&  718&    12&   31&  0.507& 0.289&   36.597 \\ 
\bottomrule
\end{tabular}}
\caption{Language models tend to prefer outputting expressions of certain confidence, such as 0.8, and 0.9.}
\label{table: whyCE}
\vspace{-0.5em}
\end{table*}

\section{Why could CE be used as a metric?} \label{sec: CE}
As mentioned in section~\ref{main_results}, we found that Language models tend to prefer outputting expressions of certain confidence, such as 'Highly Likely', 0.8, and 0.9. In the table~\ref{table: whyCE}, we have counted the occurrence of different confidence levels for various models on different datasets to demonstrate the model's preference for certain confidence levels when using the Verb and Ling method.

We also notice that as the model parameters increased, the accuracy of the model improved, but the language model's preference for certain confidence levels do not change and even became stronger. 
Therefore, we introduced the Confidence Evenness to assess whether the model's confidence is overly concentrated in certain intervals. 

Can existing metrics (such as ECE) capture this phenomenon? There is an example: on CommonSenseQA, as the parameters of Llama2-Chat increasing, the accuracy rises from 51\% to 70\%, and the ECE using the Ling method decrease from 0.385 to 0.189. But the 70B model shows a stronger preference for outputting a confidence of 0.9. Focusing solely on the ECE metric cannot fully observe the changes in model preferences. Fortunately, this phenomenal could be reflected by the CE metrics. 

Another extreme case is if models of varying parameter sizes always output a 0.9 confidence level, and as the model size increases, the average accuracy just shifts from 70\% to 90\%, then the ECE would drop to 0. If we only use existing metrics for observation, we might conclude that the model with the largest parameters has the strongest self-awareness. However, by evaluating the CE metric across different models, we can identify a potential preference in how models express confidence. Its ECE becoming 0 might just coincidentally be because the average accuracy on a certain dataset equals the confidence level it prefers to output. Therefore, we believe the CE metric provides a new perspective for observing model confidence calibration. 

Finally, it should be noted that we believe an over-concentration of model confidence in a particular value or interval is not conducive to using model confidence as a simple metric to filter out low-confidence answers.

\section{Additional Results}
\subsection{Compared with Conformal Prediction}\label{sec: comparewithconformal}
We reproduce Conformal Prediction for RLHF-LMs~\cite{kumar2023conformal} in our dataset and setting.
Specifically, for each dataset, we select 50\% samples as the calibration set and the other samples as the test set. We also set the error rate to $\alpha=0.1$ meaning the prediction answer set has a 90\% probability of containing the correct answer. We then calculate the conformal scores in the calibration set, where the specific calculation formula is $Score=1-\max{SoftmaxScore}$. For the test set, we take the $1-\alpha$ quantile of the conformal scores from the calibration set as the threshold $q$. During the testing stage, for a given sample, it is only added to the prediction set if its generated probability is greater than or equal to $1-q$. For each sample in the prediction set, we consider its confidence to be $(1-\alpha) \cdot (SoftmaxScore)$. as shown in the following table~\ref{table: conformal}, our proposed UF Calibration still demonstrates good calibration compared to conformal prediction for RLHF-LMs. It is also important to note that conformal prediction requires a calibration set to determine a threshold to build a prediction set. However, our method is a plug-and-play approach that can accurately estimate the model's confidence without requiring any prior knowledge.

\subsection{Compared with CAPE}\label{sec: cape-enum}
We reproduced the ``ENUM'' method from CAPE~\cite{jiang2023calibrating}on our evaluated dataset. 
This method (CAPE-ENUM) calibrates the probabilities of answers by permuting the order of options, which is complementary to our method
For $N$ options, we generated all possible permutations and then randomly selected 10 permutations to reorder the options, obtaining 10 different probability distributions $P_i$. 
The temperature of the language model was set to 1.0, and the final probability distribution was $P = \frac{1}{10}\sum_{i=1}^{10} P_i$. 

The results are shown in Table~\ref{table: cape}.
From the experimental results, it can be seen that the performance of these two methods is comparable.     
However, it is important to note that \textbf{CAPE-ENUM requires knowledge of the logits generated by the model for each token}. 
For Black-Box models, multiple samples are needed to obtain $P_i$, and the overall time complexity is $\mathcal O(M \cdot K)$, where $K$ represents the number of permutations, and $M$ represents the number of samples needed to obtain a probability distribution $P_i$. 
Moreover, obtaining an accurate $P_i$, usually requires a large $M$, which also leads to an increase in computational cost.

\subsection{Candidate-Aware UF Calibration}\label{sec: candidate-aware}
For some questions like ``Which of the following answers is better?'', after replace some options with ``All other options are wrong'', the remaining options are still reasonable. For example, ``Which of the following animals has the largest volume?''. 
We find that these types of questions may appear in the ARC-Challenge.
To address issues, we propose \textbf{Candidate-Aware UF Calibration}, which will introduce all the candidate answers in the prompt when utilizing our UF Calibration, even if the currently selectable options are only a subset of these.
Therefore, the model's prompt template could be changed to: ``The question is: [current question]. Candidate answers: [all candidate answers]. From the options below, please select the option you agree with the most: [options for this round]. Answer:''.
We tested Candidate-Aware UF Calibration on three Llama2-Chat models. 
Experimental results from Table~\ref{table: candidata-aware} show that Candidate-Aware UF Calibration still demonstrates performance similar to UF Calibration. 
This also partially validates that "All of the other options are incorrect" is a valid approach for quantifying fidelity.

\begin{table*}[!t]
\centering
\small
\resizebox{\textwidth}{!}{\begin{tabular}{lcccccc}
\toprule
\textbf{Model} & \textbf{Dataset} & \textbf{Method}  & \makecell{$\mathrm{ECE}_{10}$ \scalebox{0.65}{$\downarrow$}}  & \makecell{$\mathrm{BS}$ \scalebox{0.65}{$\downarrow$}} & \makecell{$\mathrm{CE}_{10}$ \scalebox{0.65}{$\uparrow$}} & \makecell{$\mathrm{IPR}_{10}$ \scalebox{0.65}{$\downarrow$}} \\ 
\midrule
\textsc{GPT-3.5-Turbo} & MMLU & Conformal Prediction & \textbf{0.086} & 0.189 & \textbf{0.897}& 0.111 \\
                       &      & Ours                 & 0.088 & \textbf{0.170} & 0.812& \textbf{0.083} \\
                       \cmidrule(lr){3-7}
                       & TruthfulQA & Conformal Prediction& 0.115& 0.197& \textbf{0.884}& \textbf{0.028}\\
                       &            & Ours                & \textbf{0.074}& \textbf{0.153}& 0.775& 0.133\\
                       \cmidrule(lr){3-7}
                       & CommonSenseQA & Conformal Prediction& 0.079& 0.173& 0.699& 0.139\\
                       &               & Ours                & \textbf{0.073}& \textbf{0.139}& \textbf{0.812}& \textbf{0.083}\\
                       \cmidrule(lr){3-7}
                       & ARC           & Conformal Prediction& \textbf{0.039}& 0.142& \textbf{0.670}& 0.143\\
                       &               & Ours                & 0.112& \textbf{0.141}& 0.897& \textbf{0.139}\\
\cmidrule(lr){1-7}
\textsc{GPT-4-Turbo}   & MMLU & Conformal Prediction & \textbf{0.084} & 0.164 & 0.482& 0.472 \\
                       &      & Ours                 & 0.089 & \textbf{0.142} & \textbf{0.906}& \textbf{0.083}  \\
                       \cmidrule(lr){3-7}
                       & TruthfulQA & Conformal Prediction& 0.046& 0.112& 0.425& 0.222\\
                       &            & Ours                & \textbf{0.042}& \textbf{0.102}& \textbf{0.764}& \textbf{0.044}\\
                       \cmidrule(lr){3-7}
                       & CommonSenseQA & Conformal Prediction& \textbf{0.040}& \textbf{0.130}& 0.509& 0.194\\
                       &               & Ours                & 0.109& 0.134& \textbf{0.925}& \textbf{0.083}\\
                       \cmidrule(lr){3-7}
                       & ARC           & Conformal Prediction& \textbf{0.084} & \textbf{0.026} & 0.000& \textbf{0.000} \\
                       &               & Ours                & 0.127 & 0.095 & \textbf{0.757}& 0.083 \\
\cmidrule(lr){1-7}												
\textsc{Baichuan2-13B-Chat}& MMLU & Conformal Prediction & 0.130 & 0.218 & \textbf{0.888}& 0.056 \\
                           &      & Ours                 & \textbf{0.076} & \textbf{0.193} & 0.829& \textbf{0.028} \\
                           \cmidrule(lr){3-7}
                           & TruthfulQA & Conformal Prediction& 0.209& 0.239& \textbf{0.865}& 0.250\\
                           &            & Ours                & \textbf{0.080}& \textbf{0.149}& 0.704& \textbf{0.028}\\
                           \cmidrule(lr){3-7} 						
                           & CommonSenseQA & Conformal Prediction& 0.056& 0.162& 0.801& \textbf{0.056}\\
                           &               & Ours                & \textbf{0.051}& \textbf{0.153}& \textbf{0.886}& \textbf{0.056}\\
                           \cmidrule(lr){3-7} 						
                           & ARC           & Conformal Prediction& \textbf{0.061}& 0.173& \textbf{0.848}& \textbf{0.028}\\
                           &               & Ours                & 0.063& \textbf{0.166}& 0.887& \textbf{0.028}\\
\cmidrule(lr){1-7}  
\textsc{Llama2-7B-Chat}    & MMLU & Conformal Prediction & 0.253 & 0.290 & 0.864 & 0.361 \\
                           &      & Ours                 & \textbf{0.102} & \textbf{0.214} & \textbf{0.890} & \textbf{0.167} \\
                           \cmidrule(lr){3-7}
                           & TruthfulQA & Conformal Prediction& 0.353 & 0.361 & \textbf{0.825} & 0.361\\
                           &            & Ours                & \textbf{0.121} & \textbf{0.186} & 0.762 & \textbf{0.083}\\
                           \cmidrule(lr){3-7}
                           & CommonSenseQA & Conformal Prediction& 0.234 & 0.283 & 0.655 & 0.333\\
                           &               & Ours                & \textbf{0.053} & \textbf{0.181} & \textbf{0.907} & \textbf{0.167}\\
                           \cmidrule(lr){3-7}
                           & ARC           & Conformal Prediction& 0.260 & 0.308 & \textbf{0.701} & \textbf{0.083}\\
                           &               & Ours                & \textbf{0.073} & \textbf{0.204}& 0.921 & 0.111\\
\cmidrule(lr){1-7} 
\textsc{Llama2-13B-Chat}   & MMLU & Conformal Prediction & 0.279 & 0.317 & 0.740 & 0.250 \\
                           &      & Ours                 & \textbf{0.070} & \textbf{0.196} & \textbf{0.852} & \textbf{0.083} \\
                           \cmidrule(lr){3-7}
                           & TruthfulQA & Conformal Prediction& 0.429 & 0.416 & 0.728 & 0.611\\
                           &            & Ours                & \textbf{0.121} & \textbf{0.180} & \textbf{0.762} & \textbf{0.083}\\
                           \cmidrule(lr){3-7}
                           & CommonSenseQA & Conformal Prediction& 0.220 & 0.274 & 0.647 & 0.250\\
                           &               & Ours                & \textbf{0.043} & \textbf{0.166} & \textbf{0.883} & \textbf{0.111}\\
                           \cmidrule(lr){3-7}
                           & ARC           & Conformal Prediction& 0.212 & 0.260 & 0.611 & 0.361\\
                           &               & Ours                & \textbf{0.069} & \textbf{0.178} & \textbf{0.886} & \textbf{0.111}\\
\cmidrule(lr){1-7} 
\textsc{Llama2-70B-Chat}   & MMLU & Conformal Prediction & 0.260 & 0.305 & 0.592 & 0.250 \\
                           &      & Ours                 &  \textbf{0.066} & \textbf{0.189} & \textbf{0.898} & \textbf{0.083} \\
                           \cmidrule(lr){3-7}
                           & TruthfulQA & Conformal Prediction& 0.281 & 0.301 & 0.558 & 0.306\\
                           &            & Ours                & \textbf{0.093} & \textbf{0.162} & \textbf{0.804} & \textbf{0.089}\\
                           \cmidrule(lr){3-7}
                           & CommonSenseQA & Conformal Prediction& 0.156 & 0.221 & 0.479 & 0.333\\
                           &               & Ours                & \textbf{0.094} & \textbf{0.156} & \textbf{0.908} & \textbf{0.111}\\
                           \cmidrule(lr){3-7}
                           & ARC           & Conformal Prediction& 0.118 & 0.189 & 0.427 & 0.361\\
                           &               & Ours                & \textbf{0.085} & \textbf{0.154} & \textbf{0.908} & \textbf{0.111}\\
\bottomrule
\end{tabular}}
\caption{Comparing calibration results of Conformal Prediction of RLHF-LMs~\cite{kumar2023conformal} and our proposed method.}
\label{table: conformal}
\vspace{-0.5em}
\end{table*}

\begin{table*}[!t]
\centering
\small
\resizebox{\textwidth}{!}{\begin{tabular}{lcccccc}
\toprule
\textbf{Model} & \textbf{Dataset} & \textbf{Method}  & \makecell{$\mathrm{ECE}_{10}$ \scalebox{0.65}{$\downarrow$}}  & \makecell{$\mathrm{BS}$ \scalebox{0.65}{$\downarrow$}} & \makecell{$\mathrm{CE}_{10}$ \scalebox{0.65}{$\uparrow$}} & \makecell{$\mathrm{IPR}_{10}$ \scalebox{0.65}{$\downarrow$}} \\ 
\midrule		
			
\textsc{Llama2-7B-Chat} & MMLU   & CAPE-ENUM   &0.099 & 0.176& 0.815 & 0.022\\
                        &        & Ours        &0.102 & 0.214& 0.890 & 0.167\\
                       \cmidrule(lr){3-7}
                       & TruthfulQA & CAPE-ENUM&0.123&0.179 &0.691 &0.200\\
                       &            & Ours     & 0.121& 0.186& 0.762&0.083\\
                       \cmidrule(lr){3-7}
                       & CommonSenseQA & CAPE-ENUM& 0.023& 0.099&0.688 &0.000\\
                       &               & Ours     & 0.053 & 0.181 & 0.907 & 0.167\\
                       \cmidrule(lr){3-7}
                       & ARC           & CAPE-ENUM  &0.066 & 0.150 & 0.808 & 0.000\\
                       &               & Ours       &0.073 & 0.204 & 0.921 & 0.111\\
\cmidrule(lr){1-7}
\textsc{Llama2-13B-Chat} & MMLU   & CAPE-ENUM   &0.104 & 0.166& 0.786&0.000\\
                        &        & Ours        &0.070 & 0.196 & 0.852 & 0.083\\
                       \cmidrule(lr){3-7}
                       & TruthfulQA & CAPE-ENUM&0.157& 0.185&0.665&0.067\\
                       &            & Ours     &0.121 & 0.180 & 0.762 & 0.083\\
                       \cmidrule(lr){3-7}
                       & CommonSenseQA & CAPE-ENUM& 0.033& 0.095& 0.650&0.000\\
                       &               & Ours     & 0.043	& 0.166& 0.883&0.083\\
                       \cmidrule(lr){3-7}
                       & ARC           & CAPE-ENUM  & 0.060& 0.125& 0.756&0.000\\
                       &               & Ours       & 0.069& 0.178&0.886&0.111\\
\cmidrule(lr){1-7}
\textsc{Llama2-70B-Chat} & MMLU   & CAPE-ENUM  & 0.098& 0.144& 0.727&0.000\\
                        &        & Ours        & 0.066& 0.189& 0.898&0.083\\
                       \cmidrule(lr){3-7}
                       & TruthfulQA & CAPE-ENUM& 0.109& 0.135& 0.576&0.133\\
                       &            & Ours     & 0.093& 0.162& 0.804&0.089\\
                       \cmidrule(lr){3-7}
                       & CommonSenseQA & CAPE-ENUM& 0.029& 0.069& 0.550&0.022\\
                       &               & Ours     & 0.094& 0.156& 0.918&0.111\\
                       \cmidrule(lr){3-7}
                       & ARC           & CAPE-ENUM  & 0.036& 0.077& 0.621&0.000\\
                       &               & Ours       & 0.085& 0.154& 0.908&0.111\\
\bottomrule
\end{tabular}}
\caption{Comparing calibration results of CAPE~\cite{jiang2023calibrating} and our proposed method.}
\label{table: cape}
\vspace{-0.5em}
\end{table*}

\begin{table*}[!t]
\centering
\small
\resizebox{\textwidth}{!}{\begin{tabular}{lcccccc}
\toprule
\textbf{Model} & \textbf{Dataset} & \textbf{Method}  & \makecell{$\mathrm{ECE}_{10}$ \scalebox{0.65}{$\downarrow$}}  & \makecell{$\mathrm{BS}$ \scalebox{0.65}{$\downarrow$}} & \makecell{$\mathrm{CE}_{10}$ \scalebox{0.65}{$\uparrow$}} & \makecell{$\mathrm{IPR}_{10}$ \scalebox{0.65}{$\downarrow$}} \\ 
\midrule		
			
\textsc{Llama2-7B-Chat} & MMLU   & Candidate-Aware   & 0.129 & 0.233 &  0.897 & 0.083\\
                        &        & Ours        &0.102 & 0.214& 0.890 & 0.167\\
                       \cmidrule(lr){3-7}
                       & TruthfulQA &  Candidate-Aware &0.166 & 0.205 &0.769& 0.156\\
                       &            & Ours     & 0.121& 0.186& 0.762&0.083\\
                       \cmidrule(lr){3-7}
                       & CommonSenseQA &  Candidate-Aware & 0.080 & 0.188 & 0.839  & 0.083\\
                       &               & Ours     & 0.053 & 0.181 & 0.907 & 0.167\\
                       \cmidrule(lr){3-7}
                       & ARC           & Candidate-Aware  & 0.103&  0.221& 0.892  & 0.111\\
                       &               & Ours       &0.073 & 0.204 & 0.921 & 0.111\\
\cmidrule(lr){1-7}
\textsc{Llama2-13B-Chat} & MMLU   & Candidate-Aware   & 0.088& 0.200 & 0.834  & 0.139\\
                        &        & Ours        &0.070 & 0.196 & 0.852 & 0.083\\
                       \cmidrule(lr){3-7}
                       & TruthfulQA & Candidate-Aware &0.152 &  0.195&  0.795 & 0.167\\
                       &            & Ours     &0.121 & 0.180 & 0.762 & 0.083\\
                       \cmidrule(lr){3-7}
                       & CommonSenseQA & Candidate-Aware & 0.053 & 0.173 &  0.885 & 0.083\\
                       &               & Ours     & 0.043	& 0.166& 0.883&0.083\\
                       \cmidrule(lr){3-7}
                       & ARC           & Candidate-Aware   &  0.062& 0.183 & 0.882  & 0.111\\
                       &               & Ours       & 0.069& 0.178&0.886&0.111\\
\cmidrule(lr){1-7}
\textsc{Llama2-70B-Chat} & MMLU   & Candidate-Aware      & 0.131&  0.232&   0.922& 0.111\\
                        &        & Ours                  & 0.066& 0.189& 0.898&0.083\\
                       \cmidrule(lr){3-7}
                       & TruthfulQA & Candidate-Aware    & 0.212 & 0.220 & 0.869  &0.378 \\
                       &            & Ours               & 0.093& 0.162& 0.804&0.089\\
                       \cmidrule(lr){3-7}
                       & CommonSenseQA & Candidate-Aware &  0.103&  0.202&  0.884 & 0.111\\
                       &               & Ours            & 0.094& 0.156& 0.918&0.111\\
                       \cmidrule(lr){3-7}
                       & ARC           & Candidate-Aware & 0.112&  0.192& 0.849  & 0.111\\
                       &               & Ours            & 0.085& 0.154& 0.908&0.111\\
\bottomrule
\end{tabular}}
\caption{Comparing calibration results of Candidate-Aware UF Calibration and UF Calibration.}
\label{table: candidata-aware}
\vspace{-0.5em}
\end{table*}

\subsection{Brier Score}
Besides the ECE metric, the Brier Score is also commonly used as an evaluation criterion for model calibration. 
\begin{align}
    \mathrm{Brier Score} = \frac{1}{N} \sum_{t=1}^N (f_t - o_t)^2, 
\end{align}
where $f_t$ is the probability and $o_t$ is the label. Accordingly,  $f_t$  can be referred to as the model's confidence, while $o_t$ represents whether it is the correct answer (0 indicating an incorrect answer, 1 indicating a correct answer).
In Table~\ref{table: brierscore}, we present the Brier Scores of various baselines and our proposed method. It can be seen that our method still exhibits good calibration, especially for closed-source models such as \texttt{GPT-3.5-Turbo}, \texttt{GPT-4 Turbo}.

\begin{figure*}[!t]
  \centering
  \includegraphics[width=0.95\textwidth]{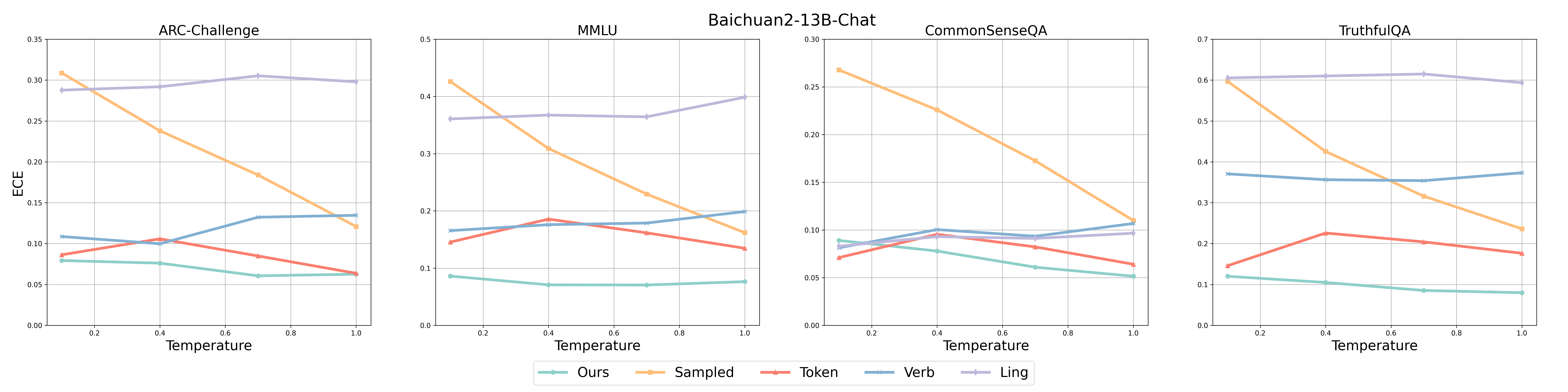}
  \vspace{-0.5em}
  \caption{The Impact of Temperature on Different Methods. Our proposed method achieved well-calibrated results across all temperatures. The experimental results are derived from \texttt{Baichuan2-13B-Chat}.} \label{fig:baichuan_temperature_scaling}
\end{figure*}

\begin{table*}[!t]
\centering
\small
\resizebox{\textwidth}{!}{\begin{tabular}{lcccccc}
\toprule
\textbf{Model} & \textbf{Method} & \textbf{ARC-Challenge} & \textbf{MMLU}  & \textbf{CommonSenseQA} & \textbf{TruthfulQA} & \textbf{Avg.} \\ 
\midrule
\textsc{GPT-3.5-Turbo} & Verb & 0.181& 0.247& 0.189 & 0.274& 0.223\\
                       & Ling & 0.197& 0.278& 0.204& 0.318& 0.249\\
                       & Sampled & 0.157& 0.202& 0.216& 0.206 & 0.195\\
                     & Conformal & \uline{0.142} & \uline{0.189} & \uline{0.173} & \uline{0.197} & \uline{0.175} \\
 & \cellcolor{gray!20} Ours & \cellcolor{gray!20} \textbf{0.141}&\cellcolor{gray!20}  \textbf{0.170}& \cellcolor{gray!20} \textbf{0.139} & \cellcolor{gray!20}\textbf{0.153} & \cellcolor{gray!20}\textbf{0.151}\\
\cmidrule(lr){1-7}
\textsc{GPT-4-Turbo} & Verb & 0.181& 0.247& 0.204& 0.274& 0.227\\
                     & Ling & 0.198& 0.278& 0.216& 0.318& 0.253\\
                     & Sampled & \uline{0.074} & 0.174 & 0.147& \uline{0.112}& 0.127\\
                     & Conformal & \textbf{0.026} & \uline{0.164} & \textbf{0.130} & \uline{0.112} & \textbf{0.108}\\
                     &\cellcolor{gray!20} Ours & \cellcolor{gray!20}0.095 & \cellcolor{gray!20}\textbf{0.142}&\cellcolor{gray!20} \uline{0.134}&\cellcolor{gray!20} \cellcolor{gray!20}\textbf{0.102}&\cellcolor{gray!20} \uline{0.118}\\
\cmidrule(lr){1-7}
\textsc{Baichuan2-13B-Chat} & Verb & 0.257& 0.294& 0.239& 0.363& 0.288\\
                            & Ling  & 0.336& 0.407& 0.235& 0.553& 0.383\\
                            & Sampled & 0.196& 0.236& 0.186& 0.262& 0.220\\
                            & Token & \textbf{0.095}& \textbf{0.168}& \textbf{0.092}& \uline{0.198} & \textbf{0.138}\\
                            & Conformal & 0.173 & {0.218} & 0.162 & 0.239 & 0.198 \\
                            & \cellcolor{gray!20} Ours & \cellcolor{gray!20}\uline{0.166}& \cellcolor{gray!20} \uline{0.193}& \cellcolor{gray!20}\uline{0.153}& \cellcolor{gray!20}\textbf{0.149}&\cellcolor{gray!20} \uline{0.165}\\
\cmidrule(lr){1-7}  
\textsc{Llama2-7B-Chat} & Verb & 0.332& 0.348& 0.283& 0.449& 0.353\\
                        & Ling & 0.451& 0.471& 0.396& 0.609& 0.4821\\
                        & Sampled & 0.358& 0.350& 0.323& 0.411& 0.360\\
                        & Token & \textbf{0.171}& \uline{0.238}& \textbf{0.158}& \uline{0.246}& \uline{0.203}\\
                        & Conformal & 0.308 & 0.290 & 0.283 & 0.361 & 0.311\\
                        &\cellcolor{gray!20} Ours &\cellcolor{gray!20} \uline{0.204}& \cellcolor{gray!20}\textbf{0.214}& \cellcolor{gray!20}\uline{0.181}& \cellcolor{gray!20}\textbf{0.186}& \cellcolor{gray!20}\textbf{0.196}\\
\cmidrule(lr){1-7} 
\textsc{Llama2-13B-Chat}& Verb & 0.277& 0.320& 0.272& 0.394& 0.316\\
                        & Ling & 0.352& 0.448& 0.343& 0.599& 0.435\\
                        & Sampled & 0.318& 0.374& 0.317& 0.470& 0.370\\
                        & Token & \textbf{0.141}& \uline{0.233}& \textbf{0.150}& \uline{0.242}& \uline{0.192}\\
                        & Conformal & 0.260 & 0.317 & 0.274 & 0.416 & 0.317\\
                        & \cellcolor{gray!20}Ours &\cellcolor{gray!20} \uline{0.178}& \cellcolor{gray!20}\textbf{0.196}& \cellcolor{gray!20}\uline{0.166}&\cellcolor{gray!20} \textbf{0.180}& \cellcolor{gray!20}\textbf{0.180}\\
\cmidrule(lr){1-7} 
\textsc{Llama2-70B-Chat} & Verb & 0.206& 0.297& 0.208& 0.332& 0.261\\
                         & Ling & 0.267& 0.390& 0.240& 0.496& 0.348\\
                         & Sampled  & 0.236& 0.347& 0.237& 0.360& 0.295\\
                         & Token & \textbf{0.094}& \uline{0.196}& \textbf{0.098}& \uline{0.174}& \textbf{0.141}\\
                         & Conformal & 0.189 & 0.305 & 0.221 & 0.301 & 0.254\\
                         & \cellcolor{gray!20}Ours & \cellcolor{gray!20}\uline{0.154} & \cellcolor{gray!20}\textbf{0.189}& \cellcolor{gray!20}\uline{0.156}&\cellcolor{gray!20} \textbf{0.162}& \cellcolor{gray!20}\uline{0.165}\\
\bottomrule
\end{tabular}}
\caption{The Brier Score of different methods from six RLHF-Models on four MCQA datasets.}
\label{table: brierscore}
\vspace{-0.5em}
\end{table*}

\section{Prompt Templates} \label{sec: prompt}
We use the prompt template from ~\citet{tian-etal-2023-just} for a fair comparison. 
The prompt template for each baseline is provided in Table~\ref{tab:prompts}. 
The question is substituted for the variable \texttt{\$\{THE\_QUESTION\}} in each prompt. 
Table~\ref{tab:linguistic} shows the linguistic expression list of confidence we used for the Ling Method, which originates from~\citet{fagen2023perception}.

\begin{table*}[!t]
    \centering
    \small
    \begin{tabular}{lp{10cm}}
        \toprule
        \textbf{Method} & \textbf{Prompt Template} \\
        \midrule
         
         Verb~\cite{tian-etal-2023-just} & \texttt{Provide your best guess and the probability that it is correct (0.0 to 1.0) for the following question. Give ONLY the guess and probability, no other words or explanation. For example:\textbackslash nGuess: <most likely option, without any extra commentary whatsoever; just the option>\textbackslash nProbability: <the probability between 0.0 and 1.0 that your guess is correct, without any extra commentary whatsoever; just the probability!>\textbackslash nThe question is: \{question\}\textbackslash nOptions:\textbackslash n\{choices\}Answer: }  \\
         \midrule
         Ling~\cite{tian-etal-2023-just} & \texttt{Provide your best guess for the following question, and describe how likely it is that your guess is correct as one of the following expressions: \{EXPRESSION\_LIST\}. Give ONLY the guess and your confidence, no other words or explanation. For example:\textbackslash n\textbackslash n Guess: <most likely guess, as short as possible; not a complete sentence, just the guess!>\textbackslash n Confidence: <description of confidence, without any extra commentary whatsoever; just a short phrase!>\textbackslash n The question is: \{question\}\textbackslash n Options:\textbackslash n\{choices\}Answer: } \\
         \midrule
         Sampled & \texttt{Provide the option you agree with most for the following question. Give ONLY the option of the answer, no other words or explanation. For example:\textbackslash nAnswer: <most likely option, without any extra commentary whatsoever; just the option>\textbackslash nThe question is: \{question\}\textbackslash nOptions:\textbackslash n\{choices\}Answer: } \\
         \midrule
         Token & \texttt{Provide the option you agree with most for the following question. Give ONLY the option of the answer, no other words or explanation. For example:\textbackslash nAnswer: <most likely option, without any extra commentary whatsoever; just the option>\textbackslash nThe question is: \{question\}\textbackslash nOptions:\textbackslash n\{choices\}Answer: } \\
         \midrule
         Ours & \texttt{Provide the option you agree with most for the following question. Give ONLY the option of the answer, no other words or explanation. For example:\textbackslash nAnswer: <most likely option, without any extra commentary whatsoever; just the option>\textbackslash nThe question is: \{question\}\textbackslash nOptions:\textbackslash n\{choices\}Answer: } \\
    \bottomrule
    \end{tabular}
    \caption{Prompt templates for each method evaluated.}
    \label{tab:prompts}
\end{table*}

\begin{table*}[h]
\centering
\resizebox{1.0\columnwidth}{!}{
\begin{small}
\begin{tabular}{cc}
\toprule
\bf Linguistic Expression &\bf Confidence Score \\

\cmidrule(lr){1-2}
    \texttt{`Certain'} & \texttt{1.0} \\
    \texttt{`Almost Certain'} & \texttt{0.95} \\
    \texttt{`Highly Likely'} & \texttt{0.9} \\
    \texttt{`Very Good Chance'} & \texttt{0.8} \\
    \texttt{`We Believe'} & \texttt{0.75} \\
    \texttt{`Probably'} & \texttt{0.7} \\
    \texttt{`Probable'} & \texttt{0.7} \\
    \texttt{`Likely'} & \texttt{0.7} \\
    \texttt{`Better than Even'} & \texttt{0.6} \\
    \texttt{`About Even'} & \texttt{0.5} \\
    \texttt{`Probably Not'} & \texttt{0.25} \\
    \texttt{`We Doubt'} & \texttt{0.2} \\
    \texttt{`Unlikely'} & \texttt{0.2} \\
    \texttt{`Little Chance'} & \texttt{0.1} \\
    \texttt{`Chances are Slight'} & \texttt{0.1} \\
    \texttt{`Improbable'} & \texttt{0.1} \\
    \texttt{`Highly Unlikely'} & \texttt{0.05} \\
    \texttt{`Almost No Chance'} & \texttt{0.02} \\
    \texttt{`Impossible'} & \texttt{0.0} \\
\bottomrule
\end{tabular}
\end{small}
 }
\caption{The \texttt{EXPRESSION\_LIST} we used for the Ling Method.}
\label{tab:linguistic}
\end{table*}

\section{Reliability Diagram} \label{sec: diagram}
\begin{figure*}[h]
  \centering
  \includegraphics[width=0.97\textwidth]{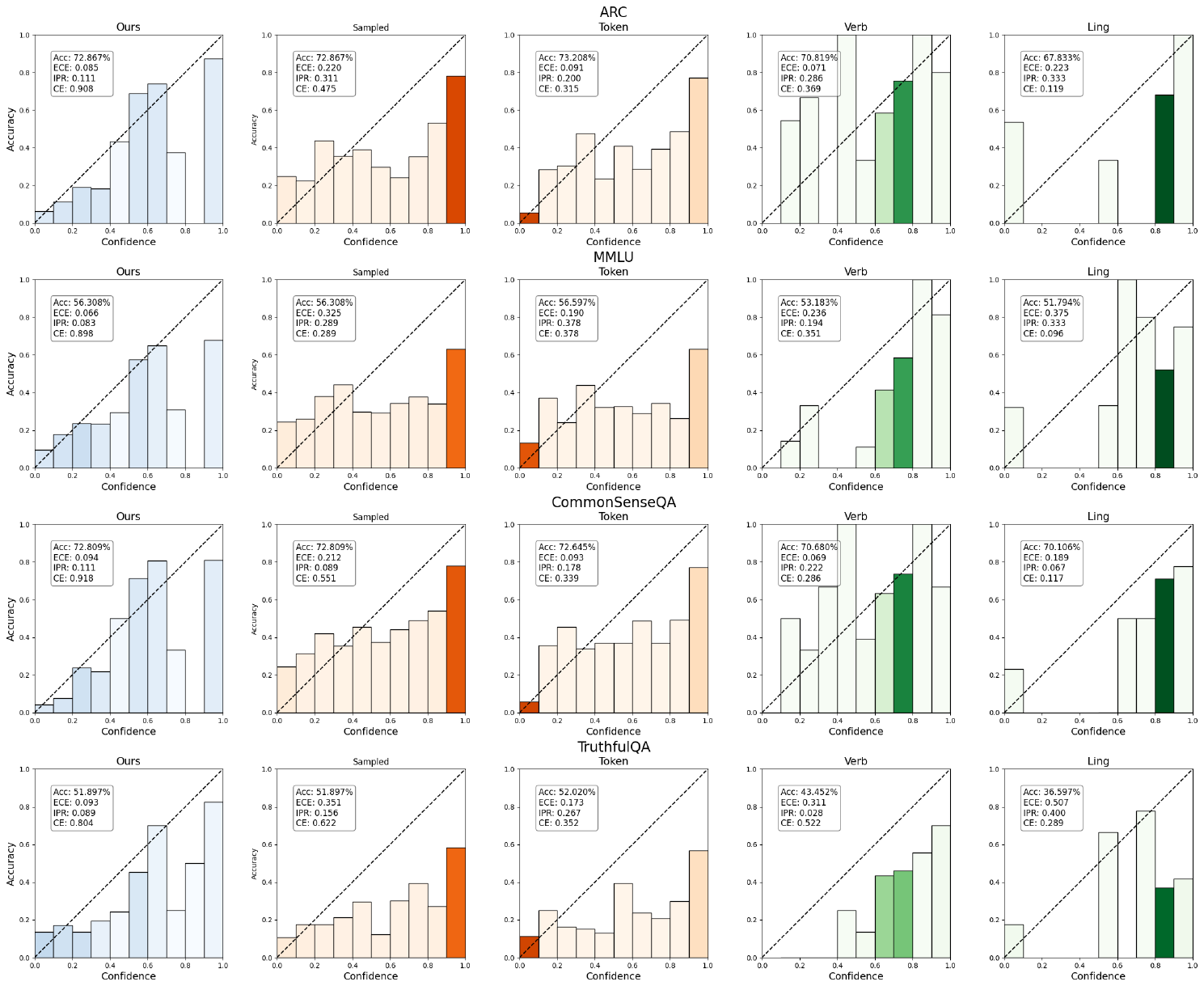}
  \vspace{-0.5em}
  \caption{The experimental results are derived from \texttt{LLaMA2-70B-Chat} on 4 MCQA datasets.} \label{fig:full-llama2_70b_chat}
\end{figure*}

\begin{figure*}[h]
  \centering
  \includegraphics[width=0.97\textwidth]{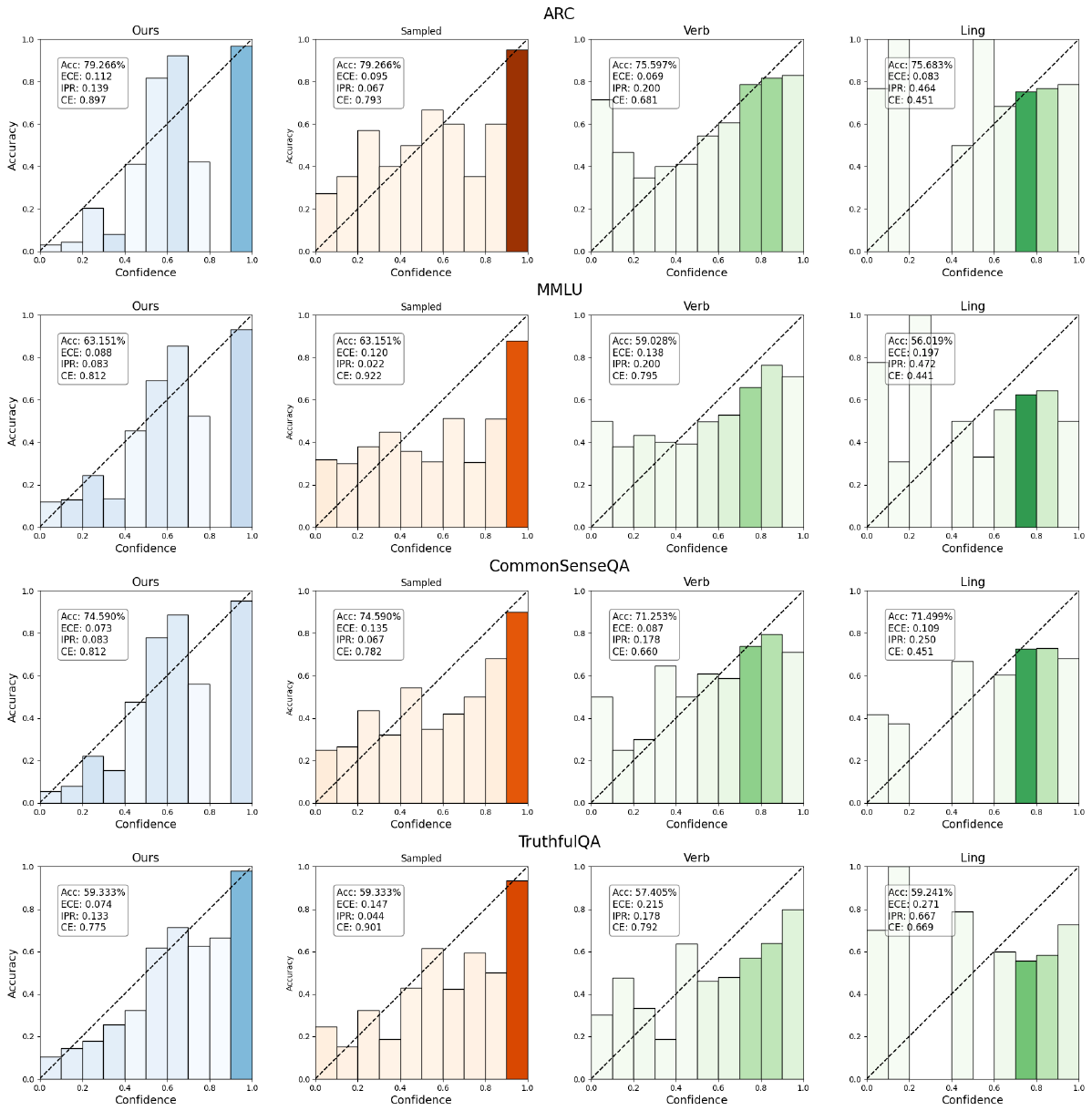}
  \vspace{-0.5em}
  \caption{The experimental results are derived from \texttt{GPT-3.5-Turbo} on 4 MCQA datasets.} \label{fig:full-gpt_3.5_turbo}
\end{figure*}

We provide the reliability diagrams of all the RLHF-LMs we evaluated in Figures~\ref{fig:full-gpt_3.5_turbo}-\ref{fig:full-llama2_70b_chat}.
In a reliability diagram, the darker the color of the bar, the greater its density is, which indicates a preference for the confidence the language models express.
Although the average accuracy of various RLHF-LMs is quite different, these model always prefer to express their confidence about 70-90\% in verbalized methods.

\clearpage
\begin{figure*}[h]
  \centering
  \includegraphics[width=0.97\textwidth]{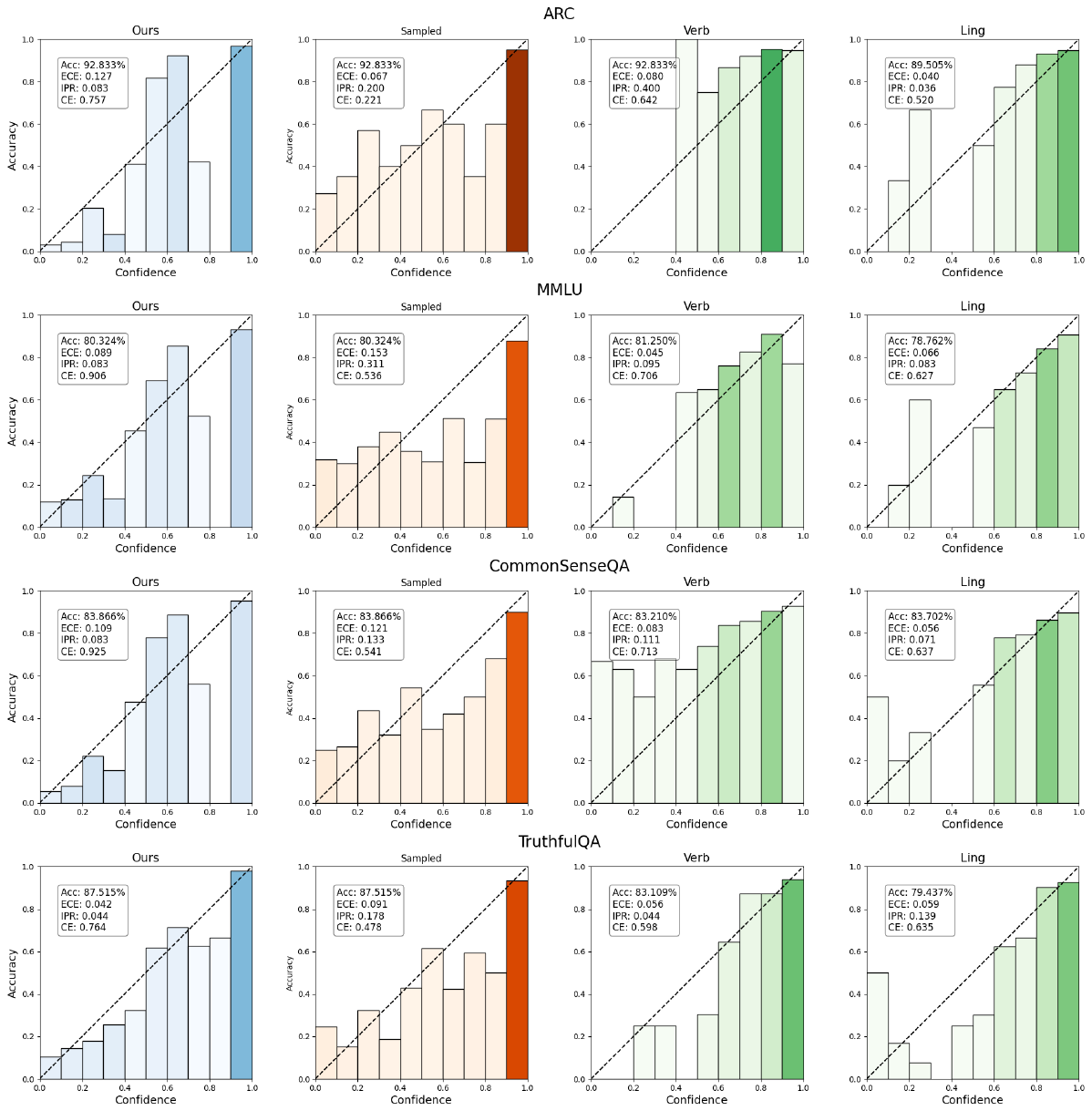}
  \vspace{-0.5em}
  \caption{The experimental results are derived from \texttt{GPT-4-Turbo} on 4 MCQA datasets.} \label{fig:full-gpt_4_turbo}
\end{figure*}

\begin{figure*}[h]
  \centering
  \includegraphics[width=0.97\textwidth]{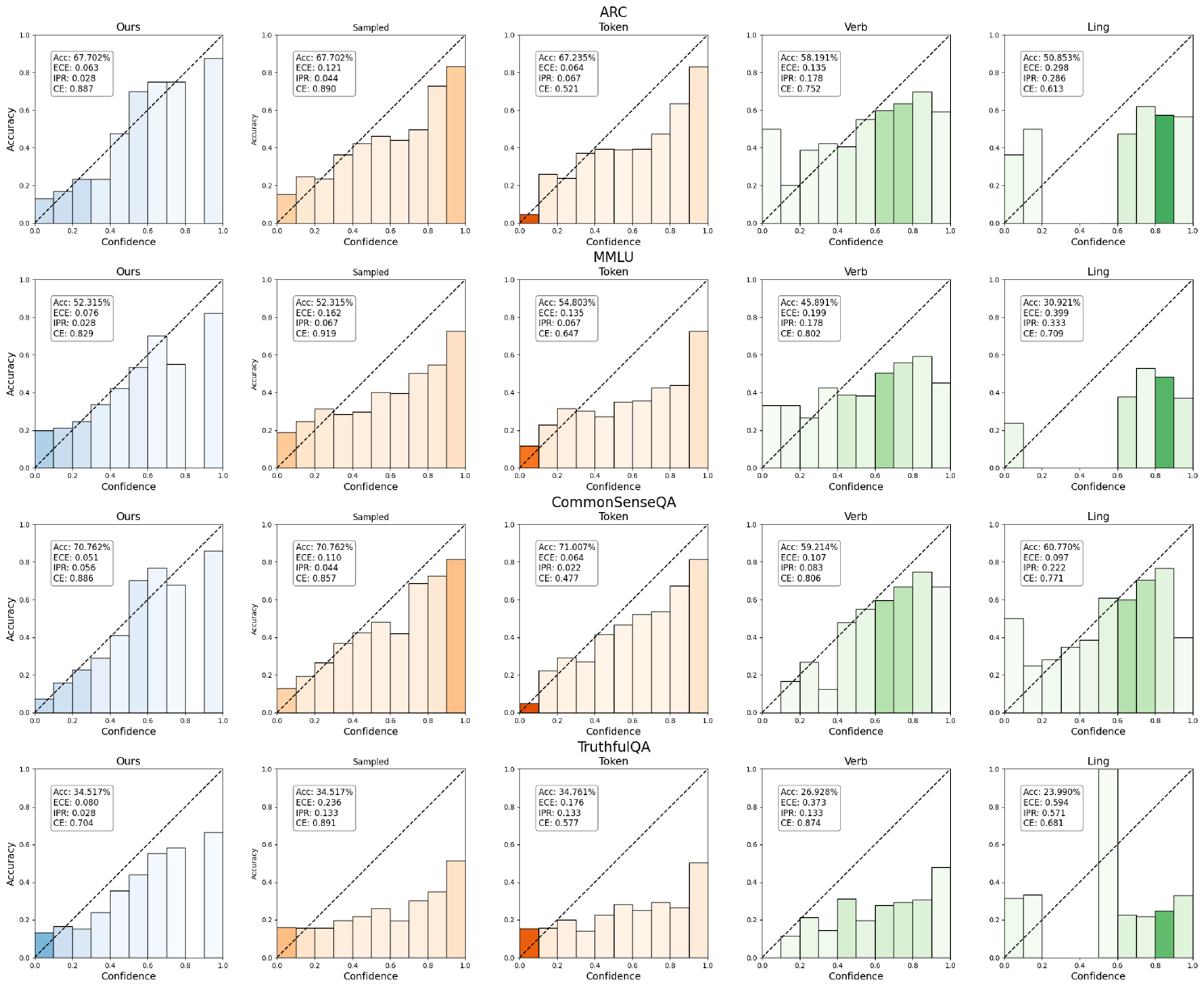}
  \vspace{-0.5em}
  \caption{The experimental results are derived from \texttt{Baichuan2-13B-Chat} on 4 MCQA datasets.} \label{fig:full-baichuan2_13b_chat}
\end{figure*}

\begin{figure*}[h]
  \centering
  \includegraphics[width=0.97\textwidth]{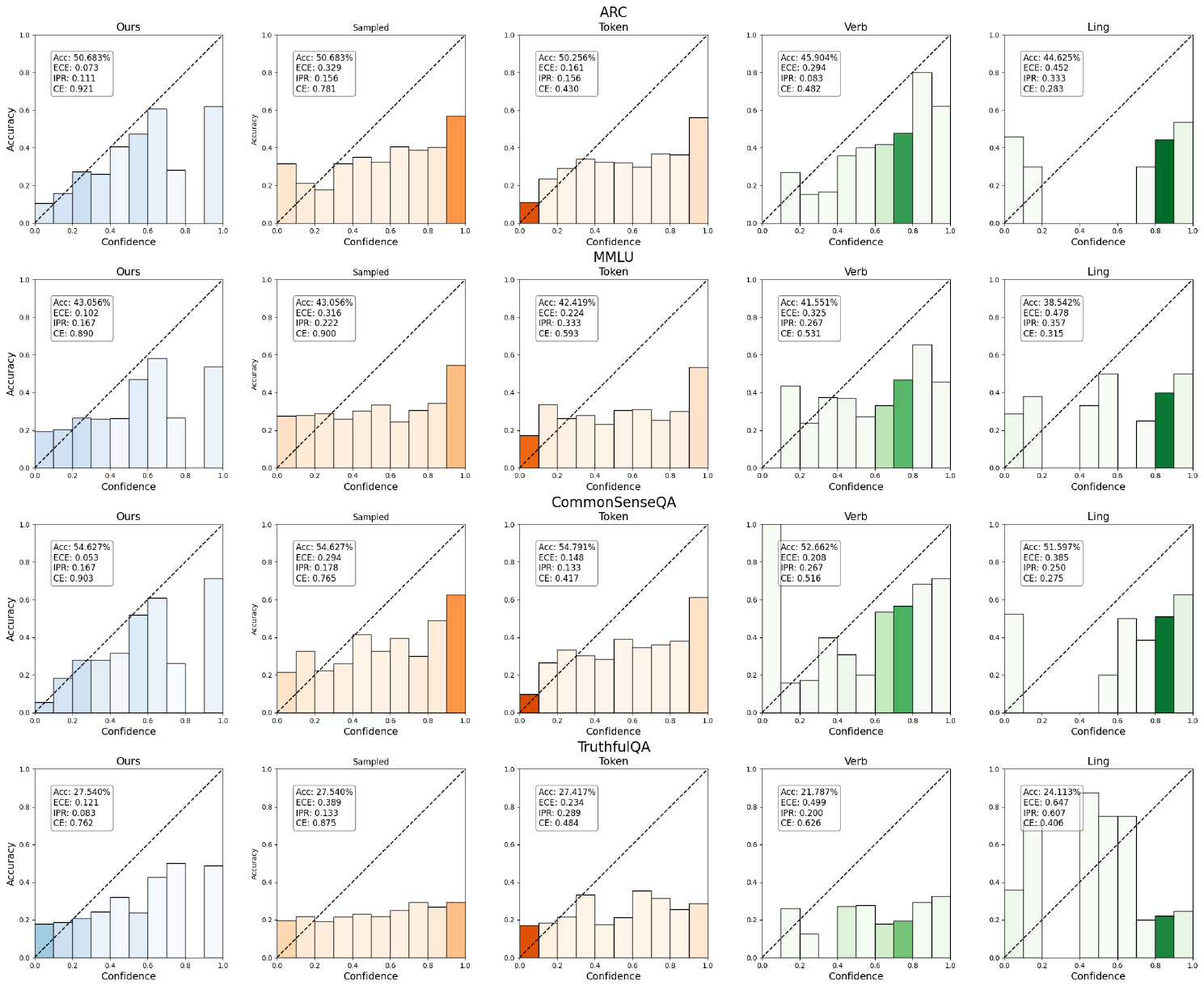}
  \vspace{-0.5em}
  \caption{The experimental results are derived from \texttt{LLaMA2-7B-Chat} on 4 MCQA datasets.} \label{fig:full-llama2_7b_chat}
\end{figure*}

\begin{figure*}[h]
  \centering
  \includegraphics[width=0.97\textwidth]{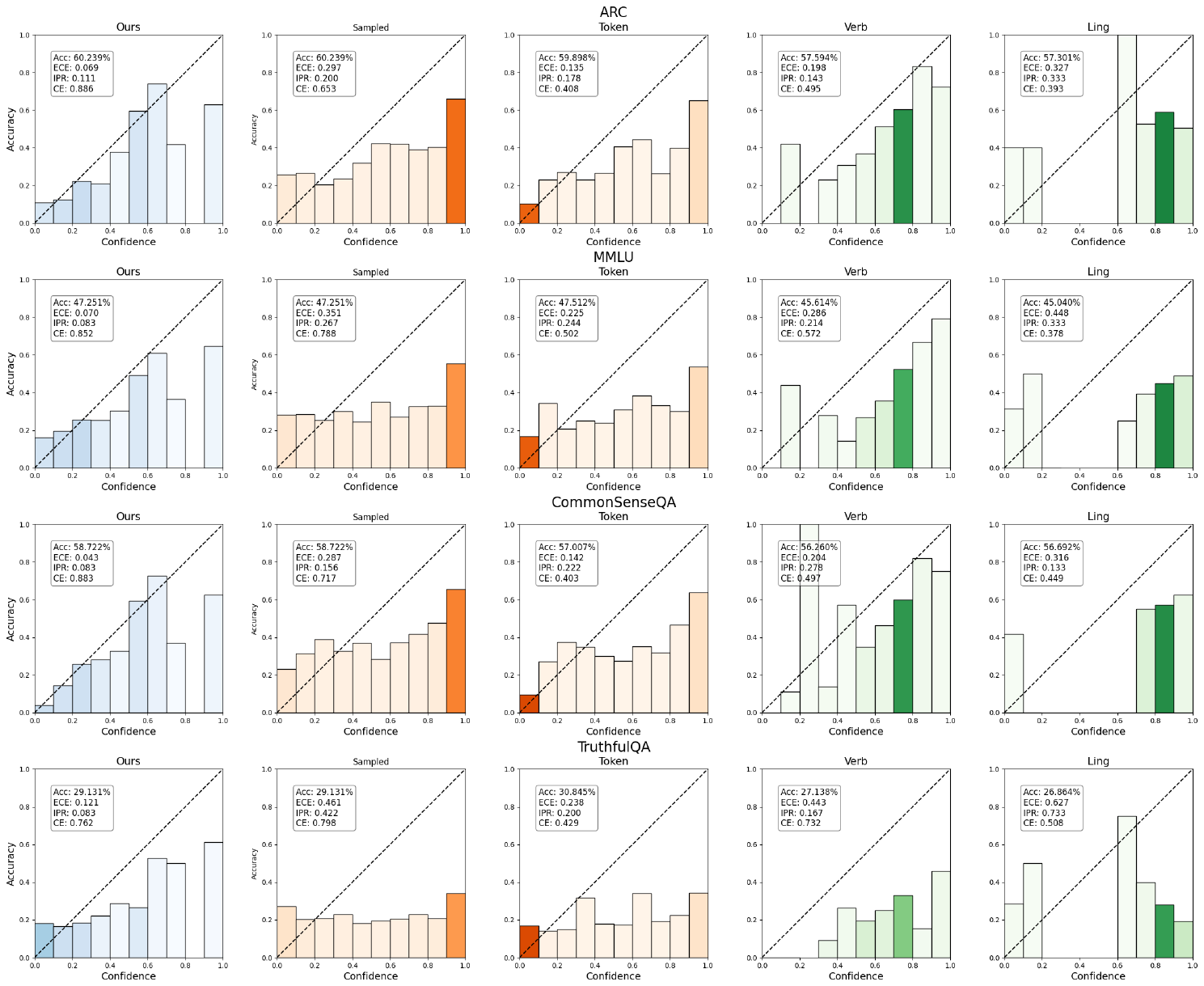}
  \vspace{-0.5em}
  \caption{The experimental results are derived from \texttt{LLaMA2-13B-Chat} on 4 MCQA datasets.} \label{fig:full-llama2_13b_chat}
\end{figure*}

\end{document}